\documentclass[10pt,twocolumn,letterpaper]{article}

\usepackage[pagenumbers]{cvpr} 
\usepackage{algorithm}
\usepackage{algorithmic}
\usepackage{multirow}
\usepackage[inline]{enumitem}
\usepackage{enumitem}
\usepackage{boldline}
\usepackage{hhline}
\usepackage{amsmath}

\usepackage{xcolor}
\usepackage{tikz}
\usepackage{colortbl}
\usepackage{etoolbox}
\usepackage[normalem]{ulem} 
\usepackage{xparse}         
\usepackage{xstring}
\definecolor{heatcolor}{HTML}{87bde7}

\newcommand{\columnheat}[3]{
    \pgfmathsetmacro{\relativepos}{100*(#1-#2)/(#3-#2)}  
    \pgfmathsetmacro{\intensity}{max(15, min(100, \relativepos))}  
    \makebox[2em]{\colorbox{heatcolor!\intensity!white}{{\makebox[3em]{#1}}}}%
}

\definecolor{cvprblue}{rgb}{0.21,0.49,0.74}
\usepackage[pagebackref,breaklinks,colorlinks,allcolors=cvprblue]{hyperref}

\def\paperID{11038} 
\def\confName{CVPR}
\def\confYear{2026}

\title{Parameter Importance-Driven Continual Learning for Foundation Models}

\author{
Lingxiang Wang\textsuperscript{1,2} \quad
Hainan Zhang\textsuperscript{1,2}\thanks{Corresponding author} \quad
Zhiming Zheng\textsuperscript{1,2} \\
\\
\textsuperscript{1}Beijing Advanced Innovation Center for Future Blockchain and Privacy Computing, Beihang University \\
\textsuperscript{2}School of Artificial Intelligence, Beihang University \\
{\tt\small wanglingxiang@buaa.edu.cn, zhanghainan@buaa.edu.cn, zhengzhiming0130@163.com}
}

\begin{document}
\maketitle
\begin{abstract}
Domain-specific post-training often causes catastrophic forgetting, making foundation models lose their general reasoning ability and limiting their adaptability to dynamic real-world environments. Preserving general capabilities while acquiring downstream domain knowledge is a central challenge for large language and multimodal models. Traditional continual learning methods, such as regularization, replay and architectural isolation, suffer from poor downstream performance, reliance on inaccessible historical data, or additional parameter overhead. While recent parameter-efficient tuning (PET) methods can alleviate forgetting, their effectiveness strongly depends on the choice of parameters and update strategies. In this paper, 
we introduce PIECE, a Parameter Importance Estimation-based Continual Enhancement method that preserves general ability while efficiently learning domain knowledge without accessing prior training data or increasing model parameters. PIECE selectively updates only 0.1\% of core parameters most relevant to new tasks, guided by two importance estimators: PIECE-F based on Fisher Information, and PIECE-S based on a second-order normalization that combines gradient and curvature information. Experiments across three language models and two multimodal models show that PIECE maintains general capabilities and achieves state-of-the-art continual learning performance across diverse downstream tasks. Our results highlight a practical path to scalable, domain-adaptive foundation models without catastrophic forgetting.
\end{abstract}
    
\section{Introduction}
\label{sec:intro}
\begin{figure}
    \centering
    \includegraphics[width=\linewidth]{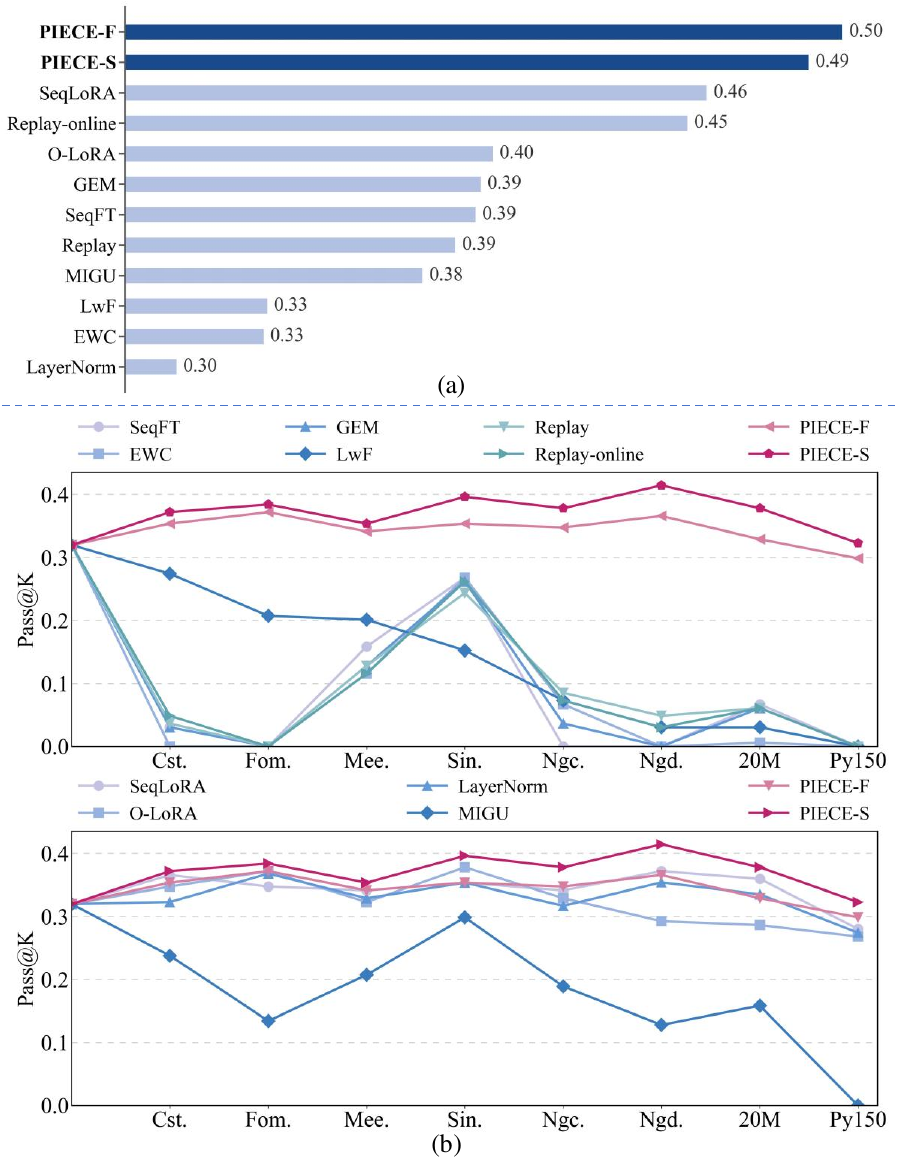}
    \vspace{-8mm}
    \caption{(a) Average downstream-task scores and (b) HumanEval Programming ability (Pass@K, K=1) of Llama3-8B on the TRACE benchmark as continual learning tasks increase. PIECE consistently outperforms full fine-tuning (SeqFT), regularization (EWC, GEM, LwF), replay (Replay, Replay-online), and PET (SeqLoRA, O-LoRA, LayerNorm, MIGU) baselines in both downstream performance and capability retention.
    }
    \label{fig:motivation}
    \vspace{-5mm}
\end{figure}
Foundation models, including large language models (LLMs) and multimodal LLMs (MLLMs), have shown remarkable generalization and reasoning abilities across diverse domains, from natural language understanding~\cite{karanikolas2023large, zhuang2022acil} and multimodal reasoning~\cite{li2025survey, mitra2024compositional} to applications in healthcare~\cite{chen2025mimo} and science~\cite{yan2025position}.
However, when fine-tuned on new domains, these models often suffer from catastrophic forgetting~\cite{french1999catastrophic,liu2020learning}, which erodes their core strengths, such as general reasoning abilities and broad knowledge, and ultimately limiting their effectiveness in domain-specific adaptation~\cite{luo2025empirical, liu2024more}.

Continual learning~\cite{mccloskey1989catastrophic, shi2024continual} has been recognized as an effective mechanism for enabling models to incrementally acquire new domain knowledge. \textbf{Regularization methods}~\cite{lopez2017gradient,kirkpatrick2017overcoming} constrain model updates to remain close to the original parameters, but it may hinder efficient learning for downstream tasks. \textbf{Replay methods}~\cite{rebuffi2017icarl,rolnick2019experience} rely on historical training data, which is impractical for post-training foundation models~\cite{touvron2023llama} and inefficient due to large-scale data replay. \textbf{Architectural isolation methods}~\cite{tong2025analytic,zhao2024sapt} mitigate forgetting by separating different tasks through mechanisms such as gating, but they introduce additional parameter overhead and impede cross-task generalization~\cite{verma2021efficient}. To compare these methods, we take programming ability as a representative proxy for the general capabilities of foundation models, because it is particularly vulnerable to destruction~\cite{chen2025towards}. Figure~\ref{fig:motivation} (b) evaluate the forgetting behaviors of these traditional approaches, showing that both Regularization and Replay methods\footnote{We only focus on continue learning without parameter growth, thus we don't compare the architectural isolation methods.} fail to meet the crucial requirement of preserving the broad and general capabilities of foundation models during continual learning.

Recently, parameter-efficient tuning (PET) methods~\cite{du2024unlocking,wang2023orthogonal,hulora, zhao2024tuning} have demonstrated the potential for anti-forgetting learning~\cite{sung2021training,xu2021raise} without relying on historical training data, task labels, or modifying model architecture. As shown in Figure~\ref{fig:motivation}, our analysis reveals that: \textbf{(1) PET outperforms traditional continual learning approaches in preserving the original capabilities of the model.} In contrast, regularization and replay methods struggle with catastrophic forgetting due to the lack of access to historical training data. \textbf{(2) Different PET strategies exhibit varying levels of general capability preservation.} SeqLoRA~\cite{hulora}, which updates only Attention components, outperforms LayerNorm~\cite{zhao2024tuning} updates on downstream performance, suggesting that domain knowledge is possibly concentrated in Attention components~\cite{liao2025data}. However, it slightly weakens original abilities due to all layers structural disruption (see Section~\ref{sec:analysis}). Thus, the choice of parameters to fine-tune and the update strategy play critical roles in domain transfer and prior knowledge retention. 

In this paper, we propose PIECE, a Parameter Importance Estimation-based Continual Enhancement method that effectively learns domain-specific knowledge while preserving general capabilities. \textbf{PIECE focuses on the internal model updates without the need for historical data or additional parameters.} It selectively updates only the most relevant 0.1\% of core parameters for new tasks. Specifically, before fine-tuning, PIECE evaluates parameter importance using two independent strategies: one based on Fisher information (PIECE-F), which measures parameter sensitivity to the current task loss, and the other based on a second-order normalization method (PIECE-S), which combines gradient and curvature information to assess parameter importance. Notably, PIECE-S is our first theory-grounded variant and better preserves original model capabilities than PIECE-F. Based on these importance scores, only the Top-0.1\% parameters are updated during fine-tuning, while the remaining parameters are frozen. This selective updating enables efficient domain adaptation while preserving the model’s original capability.

We conducted extensive experiments\footnote{See code, dataset, logs and appendix in Supplemental Materials and https://github.com/wanglingxiang0717/PIECE.} on three LLMs and two MLLMs, encompassing model sizes ranging from 2B to 14B. The results demonstrate that PIECE effectively preserves prior knowledge across diverse downstream tasks while achieving state-of-the-art continual learning performance, surpassing strong baselines. Our analysis reveals that PIECE selectively updates portions of Attention as well as LayerNorm modules at lower and deeper layers, enabling it to maintain general capabilities while efficiently assimilating new knowledge. These findings highlight a practical and scalable pathway toward domain-adaptive foundation models that mitigate catastrophic forgetting. Our contributions can be summarized as follows:

\begin{itemize}
    \item We find that parameter selection is an effective strategy to mitigate catastrophic forgetting, enabling continual learning through internal model updates without the need for historical data or additional parameters.
        
    \item We propose two parameter importance estimation methods: PIECE-F and PIECE-S. Notably, PIECE-S is our first theory-grounded variant and better preserves original model capabilities than PIECE-F.
    
    \item PIECE is a model-agnostic, scalable method that enables sustainable continual learning while preserving pre-trained abilities, as validated by extensive experiments across diverse LLMs and MLLMs.

\end{itemize}
\section{Related Work}
\label{sec:related}
\textbf{Regularization-based methods} mitigate catastrophic forgetting by constraining changes to weights or units associated with previous tasks. Some research focuses on preserving task-relevant parameters~\cite{kirkpatrick2017overcoming, aljundi2018memory}, while others emphasize constraints on the model’s intermediate or final outputs~\cite{li2017learning}. \textbf{Replay-based methods} introduce the fixed memory to store either real samples~\cite{rolnick2019experience, zhuo2023continual} or pseudo-generative examples~\cite{shin2017continual, maekawa2023generative} from previous tasks, and replay them during training to preserve the acquired knowledge. Although these two types of methods have shown significant effectiveness in mitigating catastrophic forgetting, access to pretraining data is often limited in foundation models for ordinary users, which constrains their practical applicability. \textbf{Architecture-based methods} expand or isolate model parameters to retain new knowledge while preventing the forgetting of prior knowledge. Inspired by~\citet{aljundi2017expert}, some methods~\cite{le2024mixture, chen2024low} employ gating mechanisms to assign independent modules to different tasks, thereby achieving knowledge isolation. \citet{tong2025analytic} further optimize the training of the gates to improve classification accuracy and overall model performance, while~\citet{zhao2024sapt} enhance the design of the gates to increase the utilization efficiency of task-specific modules, improving knowledge transfer. However, these methods cause the parameter size to grow linearly with the number of tasks, resulting in higher computational and storage costs. 
\textbf{Parameter-Efficient Tuning} (PET)~\cite{zhang2025parameter, ding2022delta} has become an active research direction in post-training of large models. 
Some research has shown that parameter-efficient tuning offers significant advantages in mitigating catastrophic forgetting~\cite{biderman2024lora, bhat2025parameter}, with LoRA~\cite{hulora} demonstrating particularly strong performance~\cite{zhao2024sapt}. \citet{wang2023orthogonal} proposed O-LoRA, a method that updates LoRA weights within an orthogonal subspace to enhance the model’s continual learning capability. Compared to these structured methods, unstructured parameter-efficient tuning focuses more on the model’s transfer performance on downstream tasks. \citet{sung2021training} and \citet{xu2021raise} demonstrated that tuning only a subset of task-relevant parameters can achieve better transfer performance than full fine-tuning. Subsequently, \citet{du2024unlocking} introduce MIGU, which uses unstructured parameter-efficient tuning for continual learning to retain model capabilities without historical data. However, its parameter importance relies solely on the L1-norm of outputs. Building on this, we explore unstructured parameter-efficient tuning and propose two improved parameter importance evaluation strategy.

\section{Method}
\label{sec:method}
\begin{figure}[!t]
    \centering
    \includegraphics[width=\linewidth]{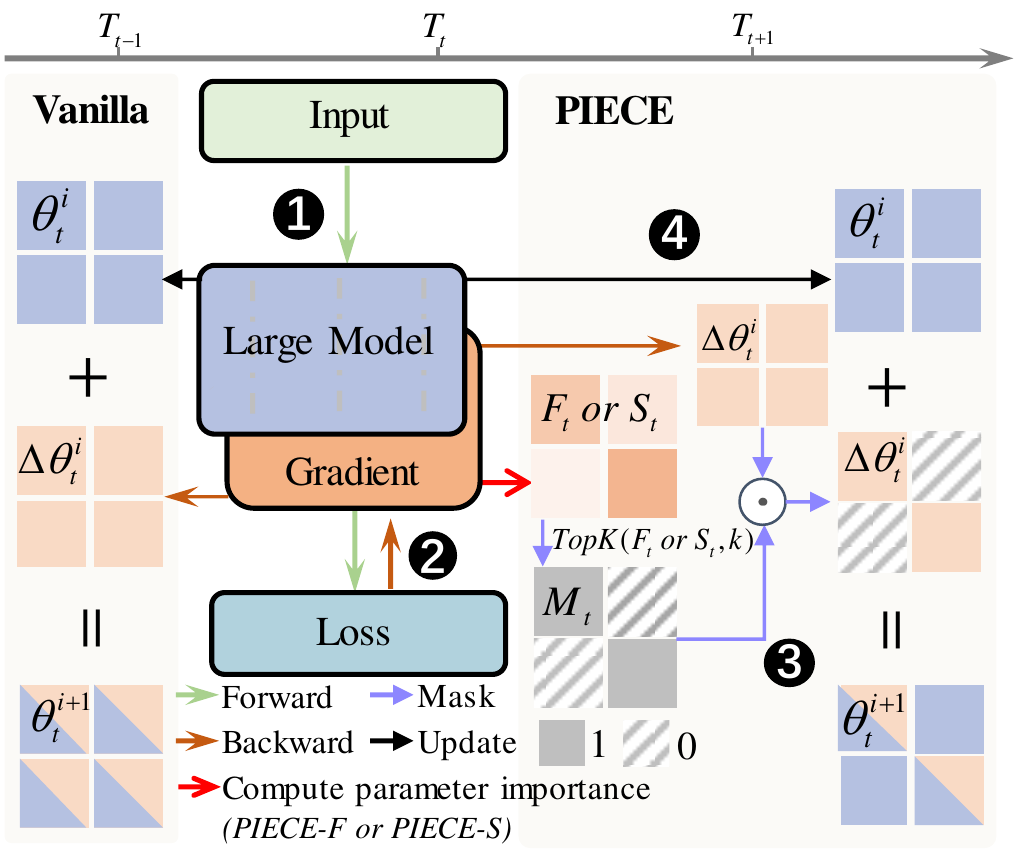}
    \caption{The illustration of PIECE.
During parameter updates, PIECE performs standard \textcircled{1} forward and \textcircled{2} backward steps, but before \textcircled{4} updating, it \textcircled{3} applies a fixed gradient mask (computed from PIECE-F/S parameter importance) to protect most parameters and update only top-k task-relevant ones.}
    \label{fig:framework}
    \vspace{-5mm}
\end{figure}
In this section, we introduce the components of PIECE, including preliminaries, two types of parameter importance estimators (PIECE-F and PIECE-S) and mask-based implementations. As shown in Figure~\ref{fig:framework}, we first evaluate the importance of each parameter based on new task data to generate a fixed gradient mask. Then, we use this mask to protect the majority of model parameters and update only a small subset of parameters most relevant to new task.
\subsection{Preliminaries}

Continual learning aims to address a key challenge: enabling a model to acquire new tasks sequentially while minimizing performance degradation on previously learned ones. Formally, the model needs to learn a sequence of tasks \(\mathcal{T} = \{T_1, T_2, \dots, T_T\}\). Each task \(T_t\) corresponds to a dataset \(\mathcal{D}_t = \{({x}_i^{(t)}, y_i^{(t)})\}_{i=1}^{n_t}\), consisting of \(n_t\) input–output pairs. Given a neural network \(f(\cdot;\theta)\) parameterized by \(\theta\), the objective at each task step \(t\) is to adapt to task \(T_t\) by minimizing:
\begin{equation}\theta^*=\arg\min_\theta\frac{1}{n_t}\sum_{i=1}^{n_t}L(f({x}_i^{(t)};\theta),y_i^{(t)}),\end{equation}
where \(L(\cdot, \cdot)\) denotes the loss function, e.g., cross-entropy. 

Typically, only the current dataset \(\mathcal{D}_t\) is accessible during the training of task \(T_t\). However, previous approaches often relax this constraint to varying degrees. For example, replay-based methods retain a small subset of past samples for replay or generate pseudo-samples; regularization-based methods store historical signals such as gradients or parameter importance to constrain updates; architecture-based methods adapt to new tasks by adding parameters or modifying the structure, sometimes with additional task-identification memory. In this work, we strictly follow a \textbf{no-history assumption}: while learning task \(T_t\), the model is prohibited from accessing any past data or intermediate training information, and both the parameter budget and architecture remain fixed.

Moreover, unlike conventional settings that start from task \(T_1\), we consider a more realistic large-model scenario in which the model may already exhibit strong performance on some tasks. These tasks should be treated as prior knowledge to be preserved. Thus, continual learning can \textbf{start from any step \(t\in\{1,\ldots,T\}\)} while protecting previously acquired capabilities.

\subsection{Fisher Information-based Importance}
We employ Fisher Information~\cite{fisher1925theory} to assess the importance of model parameters, which is widely used in modern machine learning for various purposes, such as protecting critical parameters of previous tasks~\cite{kirkpatrick2017overcoming}, guiding model compression and pruning~\cite{singh2020woodfisher}, and selecting target parameters for efficient fine-tuning~\cite{sung2021training, xu2021raise}. From the perspective of parameter importance, a parameter that strongly influences the model’s output will induce a significant change in the predictive distribution when updated. Let \(\theta\) denote the model parameters and \(x\) the input, with the output distribution denoted by \(p_\theta(y|x)\). When a small perturbation \(\delta\to0\) is applied to the parameters (as in the fine-tuning stage), it can be shown~\cite{martens2020new, pascanu2013revisiting} that the resulting change in the predictive distribution can be approximated as:
\begin{equation}\mathbb{E}_x[D_{\mathrm{KL}}(p_\theta(y|x)\|p_{\theta+\delta}(y|x))]\approx\delta^\top F_\theta\delta,\end{equation}
where \(F_\theta\) denotes the Fisher information matrix, which is defined as:
{\small \begin{equation}F_\theta=\mathbb{E}_x\left[\mathbb{E}_{y\sim p_\theta(y|x)}\left[\nabla_\theta\log p_\theta(y|x)\nabla_\theta\log p_\theta(y|x)^\top\right]\right].\end{equation}}

Since the true data distribution is typically unavailable and computing the full Fisher information matrix is expensive, we follow the approaches of \citet{kirkpatrick2017overcoming}, \citet{sung2021training}, and \citet{xu2021raise} to approximate the importance of each parameter using the diagonal elements of the empirical Fisher information matrix computed over the training set \(D_t\) of a given task. Specifically, for task \(T_t\), the Fisher Information parameter importance of the \(i\)-th parameter is defined as:
\begin{equation}\label{PIECE-F}F_{t,i}\approx\frac{1}{|D_t|}\sum_{(x,y)\in D_t}\left(\frac{\partial\log p_{\theta_{t-1}}(y|x)}{\partial{\theta_{t-1,i}}}\right)^2.\end{equation}
\subsection{Second-order Normalization Importance}
From a Bayesian perspective, we analyze the rationale of using the Fisher information to measure parameter importance for past tasks, taking the optimization of task \(T_2\) as an example:
\begin{equation}p(\theta_2\mid D_1,D_2)=\frac{p(D_2\mid\theta_2)p(\theta_2\mid D_1)}{p(D_2)}.\end{equation}
Here, the posterior distribution \(p(\theta_{2}\mid D_{1})\) from the previous task \(T_1\), serves as the prior when optimizing the new task, thereby protecting knowledge acquired from past parameters. Maximizing this posterior is equivalent to minimizing the new task loss \(L_2(\theta_2)\) augmented with a regularization term induced by the prior from the old task:
\begin{equation}\begin{aligned}
\theta_2 & =\arg\max_{\theta_2}\log p(\theta_2\mid D_1, D_2) \\
 & =\arg\min_{\theta_2} L_2(\theta)-\log p(\theta_2\mid D_1).
\end{aligned}\end{equation}
Then perform a second-order Taylor expansion of \(f(\theta_2)=\log p(\theta_2\mid D_1)\) around the optimum \(\theta_1\) of the previous task:
\begin{equation}f(\theta_2)\approx f(\theta_1)+J_{\theta_2,\theta_1}(\theta_2-\theta_1)+
\frac{1}{2}(\theta_2-\theta_1)^\top H_{\theta_2,\theta_1}(\theta_2-\theta_1),\end{equation}
\[J_{\theta_2,\theta_1}=\left.\frac{\partial f(\theta_2)}{\partial\theta_2}\right|_{\theta_2 = \theta_1},H_{\theta_{2},\theta_{1}}=\left.\frac{\partial^{2}f(\theta_2)}{\partial\theta_2^{2}}\right|_{\theta_2 =\theta_{1}}.\]
Since \(\theta_i\) is a stationary point for the previous task, the Jacobian vanishes, i.e., \(J_{\theta_2,\theta_1}=0\). Moreover, the negative expected Hessian, i.e., \(H_{\theta_{2},\theta_{1}}\)  equals the Fisher information matrix. Thus, using Fisher information to quantify parameter importance is consistent with Bayesian formulation.

Although the Fisher information characterizes parameter importance around the optimum of the previous task, it relies on the assumption that the current parameters remain close to that optimum, where the gradients vanish. However, when learning a new task \(T_t\), the initialization \(\theta_{t-1}\) is in general not a stationary point under the new data distribution, i.e., \(\nabla_{\theta_{t-1}}\log p_{\theta_{t-1}}(D_{t})\neq0.\) Neglecting the first-order term thus ignores the primary source of parameter updates for the new task. To address this, we instead expand the new task log-likelihood \(f_t^*(\theta_t)=\log p(\theta_t\mid D_t)\) at the current parameter \(\theta_{t-1}\):
\begin{equation}
\begin{split}
f_t^*(\theta_t) \approx & f_t^*(\theta_{t-1}) + J_{\theta_t,\theta_{t-1}}(\theta_t - \theta_{t-1}) \\
&+ \frac{1}{2}(\theta_t - \theta_{t-1})^\top H_{\theta_t,\theta_{t-1}}(\theta_t - \theta_{t-1}),
\end{split}
\end{equation}
and yield the approximate optimal update: 
\begin{equation}\theta_t^\star\approx\theta_{t-1}-H_{\theta_t,\theta_{t-1}}^{-1}J_{\theta_t,\theta_{t-1}}.\end{equation}
This expression indicates that the magnitude of parameter updates is jointly determined by the first-order gradient and the second-order curvature: parameters with large gradients and low curvature tend to change more during learning and thus contribute more to adaptation. Further, assuming that the parameters are near a local optimum, the Laplace approximation gives \(\theta_t^\star\sim\mathcal{N}(\theta_t^\star,H_{\theta_t^\star,\theta_{t-1}}^{-1}),\) while the current parameters \(\theta_{t-1}\) can be viewed as a degenerate distribution \(\delta(\theta_{t-1}).\) We therefore use the standardized mean shift between these two distributions as the importance measure for each parameter. To avoid computing true data expectations and Hessians, we approximate the Jacobian by the empirical gradients and the negative Hessian by the empirical Fisher information computed at \(\theta_{t-1}\) under task \(T_t\). Specifically, for task \(T_t\) the Second-order Normalized parameter importance of the \(i\)-th parameter is defined as:
\begin{equation}\label{PIECE-S}S_{t,i}=\frac{\left|\frac{1}{|D_t|}\sum_{(x,y)\in D_t}\frac{\partial\log p_{\theta_{t-1}}(y|x)}{\partial\theta_{t-1,i}}\right|}{\sqrt{\frac{1}{|D_t|}\sum_{(x,y)\in D_t}\left(\frac{\partial\log p_{\theta_{t-1}}(y|x)}{\partial\theta_{t-1,i}}\right)^2+\xi}},\end{equation}
where \(\xi > 0\) is a constant for numerical stability.
\subsection{PIECE}
%
%
%
We introduce the core workflow of PIECE using the most standard single-step fine-tuning paradigm as an example. Let \(\theta_t^i\) denote the model parameters at the \(i\)-th iteration on task \(T_t\) (where \(\theta_t^0\) denotes the initialization for task \(T_t\)). In standard fine-tuning, gradients of the loss \(L(f(x^{(t)};\theta_t^i), y^{(t)})\) are computed and applied to update all parameters via gradient descent:
\begin{equation}\theta_{t}^{i+1}=\theta_t^{i}-\eta\frac{\partial{L}(f(x^{(t)};\theta_t^i), y^{(t)})}{\partial\theta_t^{i}}.\end{equation}
In contrast, PIECE first computes the importance of each parameter from the current task data (as defined in Eq.~\ref{PIECE-F} or Eq.~\ref{PIECE-S}), and only updates the top-\(k\) most important parameters, forming a task-specific sparse update subnetwork \(\mathcal{C}_t^k\). In practice, we set \(k = \lfloor 0.001 \cdot |\theta| \rfloor\), i.e., only \(0.1\%\) of parameters are trainable for each task. To achieve this, we define a binary mask with the same shape as \(\theta\):
\begin{equation}\left.M_{t,i}=\left\{
\begin{array}
{ll}1, & \theta_{t,i}\in\mathcal{C}_t^k \\
0, & \theta_{t,i}\notin\mathcal{C}_t^k
\end{array}\right.\right..\end{equation}
It is then applied to filter gradient updates, freezing low-importance parameters and updating the important ones:
\begin{equation}\theta_{t}^{i+1}=\theta_t^{i}-\eta(\frac{\partial{L}(f(x^{(t)};\theta_t^i), y^{(t)})}{\partial\theta_t^{i}}\odot M_{t}).\end{equation}
Algorithm~\ref{alg:piece} provides the complete pseudo-code of PIECE in a sequential multi-task learning setting.
\begin{algorithm}[]
\caption{PIECE: Parameter Importance Estimation-based Continual Enhancement}
\label{alg:piece}
\begin{algorithmic}[1]

\REQUIRE Sequential tasks $\{T_1, T_2, ..., T_T\}$; model $f(\cdot;\theta)$; top-$k$ ratio $k$; learning rate $\eta$; 
\FOR{$t = 1$ to $T$}
    \STATE Initialize $\theta_t^0 = \theta_{t-1}$ \\
        // \emph{$\theta_{0}$ is foundation model parameters.}

    \STATE Sample dataset $D_t$
    \FOR{each parameter $\theta_{t-1,i}$}
        \STATE Compute importance score $F_{t,i}$ or $S_{t,i}$ \\ via Eq.~\ref{PIECE-F} or Eq.~\ref{PIECE-S}
    \ENDFOR \\
    // \emph{Compute parameter importance for task $T_t$}

    \STATE $\mathcal{C}_t^k = \text{TopK}(F_t \enspace or \enspace S_t, k)$
    \STATE Construct binary mask $M_t$ where $M_{t,i} = 1$ if $\theta_{t,i} \in \mathcal{C}_t^k$, else $0$ \\
    // \emph{Build sparse update subnetwork} \\
    // \emph{$I_t$: number of optimization steps for task $T_t$}
    \FOR{$i = 0$ to $I_t$} 
        \STATE Compute gradient $g = \frac{\partial{L}(f(x^{(t)};\theta_t^i), y^{(t)})}{\partial\theta_t^{i}}$
        \STATE $\tilde{g} = g \odot M_t$
        \STATE $\theta_t^{i+1} = \theta_t^i - \eta \tilde{g}$
    \ENDFOR \\
    // \emph{Gradient descent on selected parameters}
\ENDFOR
\RETURN \(\theta_T\)
\end{algorithmic}
\end{algorithm}

\section{Experiments}
\label{sec:experiments}
\begin{table*}[t]
\centering
\resizebox{\textwidth}{!}{
\setlength{\tabcolsep}{1.4mm}{
\renewcommand{\arraystretch}{0.9}
\newcolumntype{C}{>{\centering\arraybackslash}p{0.96cm}}
\begin{tabular}{
    l
    !{\hspace{0.8mm}}
    *{3}{C}
    !{\hspace{0.8mm}}
    *{3}{C}
    !{\hspace{0.8mm}}
    *{3}{C}
}
\hlineB{4}
\multirow{2}{*}{Methods} & \multicolumn{3}{c}{Order 1} & \multicolumn{3}{c}{Order 2} & \multicolumn{3}{c}{Order 3} \\ 
\cmidrule(l{0pt}r{1.2em}){2-4}
\cmidrule(l{0pt}r{1.2em}){5-7}
\cmidrule(l{0pt}r{1.2em}){8-10}
& \makebox[0pt][c]{
  \scalebox{0.92}[1]{\small HumanEval}\hspace{1.2mm}
}
& \small OP
& \small BWT
& \makebox[0pt][c]{
  \scalebox{0.92}[1]{\small HumanEval}\hspace{1.2mm}
}
& \small OP
& \small BWT
& \makebox[0pt][c]{
  \scalebox{0.92}[1]{\small HumanEval}\hspace{1.2mm}
}
& \small OP
& \small BWT
\\ \hline
\multicolumn{10}{l}{
    \rule[-0.5ex]{0pt}{3.0ex}\emph{Gemma2-2B}
} \\ \hline
SeqFT           & 0.0  & 15.13  & -19.08 & 0.0 & 13.98  &	-15.47  & 0.0 & 9.19 & -21.74  \\
EWC             & 0.0  & 17.13  & -13.99 & 0.0 & 16.93  &	-10.65  & 0.0 & 9.88 & -25.38 \\
GEM             & 0.0  & 19.15  & -16.27 & 0.0 & 15.53  &	-13.32  & 0.0 & 8.47 & -23.10  \\
LwF             & 0.0  & 16.86  & -10.10 & 0.0 & 15.40  &	-4.80  & 0.0 & 5.93 & -14.48 \\
replay          & 0.0  & 26.14  & \underline{-1.43} & 0.0 & 27.60  & \textbf{0.84}  & 0.0 & 26.94 & \textbf{-0.61} \\
replay-online   & 0.0  & 29.10  & -3.82 & 0.0 & {29.50}  &	-2.18  & 0.0 & 28.38 & -4.15 \\
SeqLoRA         & 10.37 & 39.77 & -10.80 & 7.93 & 31.88  &	-7.69  & 9.15 & 29.95 & -9.17 \\
O-LoRA          & 13.41 & 43.15 & -3.45 & 12.20 & \underline{36.56}  &	-6.13  & 0.61 & 30.25 & -11.39  \\
LayerNorm       & 21.95 & 27.23 & -1.67 & \underline{19.51} & 23.38  &	\underline{-0.64}  & 20.73 & 23.04 & \underline{-0.76} \\
MIGU            & 1.22  & 22.71 & -10.36 & 0.0 & 26.18  &	-5.25  & 0.0 & 21.65 & -14.11  \\
\rowcolor{black!6}
PIECE-F (ours)    & \underline{32.32} & \textbf{44.33} & -3.21 & \underline{19.51} & 28.01  &	-5.15  & \textbf{31.71} & \underline{41.39} & -7.14  \\
\rowcolor{black!6}
PIECE-S (ours)    & \textbf{33.54} & \underline{42.03} & \textbf{-1.18} & \textbf{31.10} & \textbf{44.60}  &	-3.97  & \underline{26.22} & \textbf{42.80} & -4.26  \\
\hline
\multicolumn{10}{l}{
    \rule[-0.5ex]{0pt}{3.0ex}\emph{Llama3-8B}
} \\ \hline
SeqFT           & 0.0  & 39.22 & -10.69 & 0.61 & 29.78 & -10.25  & 0.0 & 20.10 & -28.79  \\
EWC             & 0.0  & 32.95 & -10.05 & 0.0  & 34.33 & -11.27 & 2.44 & 36.92 & -6.74 \\
GEM             & 0.0  & 39.37 & -10.24 & 0.0  & 32.03 & -12.44 & 0.0  & 39.07 & -7.15  \\
LwF             & 0.0  & 33.06 & -9.94  & 1.83 & 27.20 & -8.30  & 1.22 & 26.69 & -13.53 \\
replay          & 0.0  & 38.62 & -2.94  & 0.61 & 41.90 & \textbf{-0.24} & 0.61 &	39.56 &	-3.87 \\
replay-online   & 0.0  & 45.47 & -1.99  & 0.0  & 41.33 & -1.48  & 0.0  & 40.90 & \underline{-1.64} \\
SeqLoRA         & 28.05 & 46.04 & -4.39 & 27.44 & \underline{44.36} & -7.62 & 30.49 & 32.14 & -2.80 \\
O-LoRA          & 26.83 & 39.73 & -7.19 & 26.22 & 41.39 & -4.07 & 29.27 & {43.43} & -2.43 \\
LayerNorm       & 27.44  & 30.39 & \underline{-1.86} & \underline{34.76} & 31.01  &	\underline{-0.98}  & \underline{31.71} & 31.36 & \textbf{-0.32}  \\
MIGU            & 0.0  & 37.65 & -10.02 & 4.88 & 43.76  &	-6.21  & 1.22 & \underline{44.45} & -6.26  \\
\rowcolor{black!6}
PIECE-F (ours)    & \underline{29.88} & \textbf{50.05} & -2.66 & {31.71} & 38.54 & -3.35 & 30.49 & 36.05 & -3.11 \\
\rowcolor{black!6}
PIECE-S (ours)    & \textbf{32.32} & \underline{49.05} & \textbf{-1.78} & \textbf{35.37} & \textbf{45.55} & {-1.37}  & \textbf{34.76} & \textbf{44.79} & {-2.42} \\
\hline
\multicolumn{10}{l}{
    \rule[-0.5ex]{0pt}{3.0ex}\emph{Qwen3-14B $^*$}
} \\ \hline
SeqFT           & 57.93  & 47.50 & -6.42 & \textbf{62.80} & 43.31  &	-9.55  & \underline{58.54} & 43.97 & -10.67 \\
EWC             & 58.54  & 47.22 & -6.67 & 60.98 & 40.55  &	-12.45  & 54.88 & 47.12 & -6.51 \\
GEM             & \textbf{61.59}  & 42.73 & -8.84 & 49.39 & 43.35  &	-5.49  & 57.32 & 42.46 & -5.79 \\
LwF             & {58.54}  & 39.57 & -6.62 & 53.66 & 37.21  &	-4.19  & 54.88 & 34.46 & \underline{0.59} \\
replay          & 51.83  & 47.82 & -2.40 & \textbf{62.80} & \underline{49.96}  &	-1.08  & 53.05 & 49.37 & \textbf{1.29} \\
replay-online   & 57.32  & 49.09 & -2.60 & \underline{62.20} & 49.48  &	-0.25  & \textbf{63.41} & 49.30 & -2.50 \\
SeqLoRA         & 57.32 & 44.94 & \textbf{4.74} & 47.56 & 36.71  &	\textbf{3.71}  & 40.85 & 33.40 & -3.77 \\
O-LoRA          & 44.51 & 43.23 & -2.20 & 50.00 & 39.73  &	\underline{1.78}  & 46.95 & 31.48 & -2.98 \\
LayerNorm       & 47.56  & 22.16 & -0.17 & 50.00 & 22.35  &	0.02  & 46.95 & 22.16 & -0.19  \\
MIGU            & 42.68  & \textbf{50.95} & -0.90 & 30.49 & 43.62  &	-1.42  & 45.73 & 46.53 & -7.60  \\
\rowcolor{black!6}
PIECE-F (ours)    & 50.61 & \underline{49.80} & {-0.33} & 51.83 & {49.66}  &	-5.37  & 50.00 & \underline{51.77} & -1.04 \\
\rowcolor{black!6}
PIECE-S (ours)    & \underline{59.15} & {49.35} & \underline{-0.16} & 61.59 & \textbf{50.11}  &	0.69  & 56.10 & \textbf{53.15} & -1.61 \\
\hlineB{4}
\end{tabular}}}
\vspace{-3mm}
\caption{Performance of methods on language tasks. OP and BWT are computed on sequential tasks; inherent capability variation is measured via HumanEval. $^*$Qwen3-14B trained one epoch per task, showing less forgetting.}
\label{tab1}
\end{table*}

\begin{table}[!t]
\centering
\resizebox{\linewidth}{!}{
\setlength{\tabcolsep}{1.4mm}{
\begin{tabular}{lcccccc}
\hlineB{4}
\multirow{2}{*}{Methods} & \multirow{2}{*}{Flickr30k$(\uparrow)$} & \multicolumn{3}{c}{Various task in VQA} & \multirow{2}{*}{$OP(\uparrow)$} & \multirow{2}{*}{$BWT(\uparrow)$} \\ 
\cline{3-5}& & Act. & Com. & Cou. & \\ \hline
{\emph{Qwen3-VL-4B}} & \emph{26.19} & \multicolumn{5}{l}{}\\ \hline
SeqFT & \columnheat{7.42}{4}{40} & 47.29 & 48.66 & 44.03 & \columnheat{46.66}{34}{65} & \columnheat{-13.86}{-20}{-0.5} \\
EWC & \columnheat{4.34}{4}{40} & 40.06 & 40.42 & 38.40 & \columnheat{39.62}{34}{65} & \columnheat{-18.14}{-20}{-0.5} \\
GEM & \columnheat{8.28}{4}{40} & 47.64 & 42.54 & 41.34 & \columnheat{43.84}{34}{65} & \columnheat{-12.89}{-20}{-0.5} \\
LwF & \columnheat{26.67}{4}{40} & \textbf{66.97} & 58.19 & 40.89 & \columnheat{55.35}{34}{65} & \columnheat{-5.74}{-20}{-0.5} \\
replay & \columnheat{20.79}{4}{40} & 61.20 & 60.88 & \textbf{52.27} & \textbf{\columnheat{58.11}{34}{65}} & \columnheat{-2.72}{-20}{-0.5} \\
replay-online & \columnheat{8.86}{4}{40} & \underline{65.65} & \textbf{69.68} & \underline{50.67} & \columnheat{62.00}{34}{65} & \columnheat{-0.83}{-20}{-0.5} \\
SeqLoRA & \columnheat{32.18}{4}{40} & 38.66 & 45.56 & 23.38 & \columnheat{35.87}{34}{65} & \columnheat{-1.62}{-20}{-0.5} \\
O-LoRA & \columnheat{5.23}{4}{40} & 40.06 & 48.90 & 37.01 & \columnheat{41.99}{34}{65} & \columnheat{-1.25}{-20}{-0.5} \\
LayerNorm & \columnheat{23.77}{4}{40} & 60.29 & 59.09 & 43.06 & \columnheat{54.15}{34}{65} & \columnheat{-1.37}{-20}{-0.5} \\
MIGU & \columnheat{27.44}{4}{40} & 47.64 & 57.46 & 37.29 & \columnheat{47.46}{34}{65} & \columnheat{-4.87}{34}{65} \\
PIECE-F (ours) & \columnheat{35.59}{4}{40} & 62.66 & 58.76 & 48.19 & \columnheat{56.54}{34}{65} & \columnheat{-1.53}{-20}{-0.5} \\
PIECE-S (ours) & \textbf{\columnheat{38.74}{4}{40}} & 62.15 & \underline{61.53} & 47.50 & \columnheat{57.06}{34}{65} & \textbf{\columnheat{-0.50}{-20}{-0.5}} \\
\hline
{\emph{LLaVA-1.5-7B}} & \emph{30.82} & \multicolumn{5}{l}{}\\ \hline
SeqFT  & \columnheat{10.69}{6}{45} & 41.72 & 39.36 & 43.03 & \columnheat{41.37}{40}{68.7} & \columnheat{-26.84}{-26.9}{-0.16} \\
EWC & \columnheat{21.92}{6}{45} & 46.38 & 52.40 & 24.21 & \columnheat{41.00}{40}{68.7} & \columnheat{-10.04}{-26.9}{-0.16} \\
GEM & \columnheat{24.17}{6}{45} & 49.65 & 53.63 & 30.16 & \columnheat{44.48}{40}{68.7} & \columnheat{-7.18}{-26.9}{-0.16} \\
LwF & \columnheat{36.02}{6}{45} & 67.66 & 58.11 & 2.07 & \columnheat{42.62}{40}{68.7} & \columnheat{-0.51}{-26.9}{-0.16} \\
replay & \columnheat{13.61}{6}{45} & 65.79 & 70.90 & 45.62 & \columnheat{60.77}{40}{68.7} & \columnheat{-0.78}{-26.9}{-0.16} \\
replay-online & \columnheat{10.46}{6}{45} & 67.25 & 69.11 & 44.93 & \columnheat{60.43}{40}{68.7} & \columnheat{-1.66}{-26.9}{-0.16} \\
SeqLoRA & \columnheat{31.04}{6}{45} & \textbf{75.31} & 75.63 & \underline{53.10} & \columnheat{68.01}{40}{68.7} & \columnheat{-0.51}{-26.9}{-0.16} \\
O-LoRA & \columnheat{6.80}{6}{45} & 70.38 & 72.94 & 52.30 & \columnheat{65.21}{40}{68.7} & \columnheat{-0.26}{-26.9}{-0.16} \\
LayerNorm & \columnheat{30.71}{6}{45} & 69.68 & 60.39 & 2.91 & \columnheat{44.33}{40}{68.7} & \columnheat{-0.20}{-26.9}{-0.16} \\
MIGU & \columnheat{21.81}{6}{45} & 67.66 & 52.40 & 42.89 & \columnheat{54.32}{40}{68.7} & \columnheat{-8.62}{-26.9}{-0.16} \\
PIECE-F (ours) & \columnheat{37.51}{6}{45} & \underline{74.48} & \underline{77.18} & 53.06 & \columnheat{68.24}{40}{68.7} & \columnheat{-0.29}{-26.9}{-0.16} \\
PIECE-S (ours) & \textbf{\columnheat{41.94}{6}{45}} & \textbf{75.31} & \textbf{77.34} & \textbf{53.27} & \textbf{\columnheat{68.64}{40}{68.7}} & \textbf{\columnheat{-0.16}{-26.9}{-0.16}} \\
\hlineB{4}
\end{tabular}}}
\caption{Results of different methods on multimodal tasks.}
\label{tab2}
\vspace{-4mm}
\end{table}
We validate our method on several benchmark datasets against a variety of continual learning baselines.
\subsection{Experiment Setup}
\textbf{Datasets}
For language tasks, we use the TRACE~\cite{wang2023trace} benchmark with 1,000 training samples per task. TRACE contains eight sub-datasets spanning science (Sin.), finance (Fom.), social news (Mee.), multiple languages (English, German [20M.], Chinese [Cst.]), and complex tasks such as code questions (Py150) and mathematical reasoning (Ngc, Ngd). We use the HumanEval~\cite{chen2021evaluating} programming benchmark to assess the model’s capability retention.
For multi-modal tasks, we use VQA v2~\cite{goyal2017making}. Following VQACL~\cite{zhang2023vqacl}, tasks are split by question type with 1,000 training samples per task, keeping the original test set. We focus on three types—action (act.), commonsense (com.), and count (cou.)—based on performance gains and capability preservation. Flickr30K~\cite{young2014image} is used to assess overall capability retention. Details in supplementary materials A. \\
\textbf{Baselines} We compared three categories of baselines: \textbf{Regularization} methods like EWC~\cite{kirkpatrick2017overcoming}, GEM~\cite{lopez2017gradient}, and LwF~\cite{li2017learning} constrain updates via parameter importance, gradient projection, or distillation to preserve prior knowledge. \textbf{Replay} methods reuse past data to reduce forgetting, either after each new task (Replay)~\cite{rebuffi2017icarl} or continuously during training (Replay-online)~\cite{rolnick2019experience}. \textbf{PET}  baselines include SeqLoRA~\cite{hulora}, O-LoRA~\cite{wang2023orthogonal}, LayerNorm~\cite{zhao2024tuning}, and MIGU~\cite{du2024unlocking}, which adapt different subsets or structures of parameters. Full fine-tuning (SeqFT) is also included. Details in supplementary materials B.\\
\textbf{Metrics} Let \(R_{t,i}\) denote the model’s performance on task \(i\) after learning task \(t\). Accuracy is used for classification or single-token QA tasks, and the average of Rouge-L and BLEU scores for others. The evaluation metrics are
Overall Performance (OP)~\cite{chaudhry2018riemannian} \(OP_t = \frac{1}{t}\sum_{i=1}^t R_{t,i}\), measuring average performance across all learned tasks; Backward Transfer (BWT)~\cite{lopez2017gradient} \(BWT_t = \frac{1}{t}\sum_{i=1}^{t-1}(R_{t,i} - R_{i,i})\), indicating how later tasks affect earlier ones. These are computed on sequential tasks to compare methods and isolate changes in the model’s original abilities.
For capability preservation evaluation, code generation (HumanEval) is measured by Pass@1, i.e., the fraction of correct single-attempt outputs, and Image captioning (Flickr30K) is measured by the mean of Rouge-L and BLEU scores. See supplementary material C for details.\\
\textbf{Models and Implementation Details} We use Gemma2-2B~\cite{gemma_2024}, Llama3-8B~\cite{llama3modelcard}, and Qwen3-14B~\cite{qwen3technicalreport} for language tasks, and Qwen3-VL-4B~\cite{qwen3technicalreport} and LLaVA-1.5-7B~\cite{liu2023visual, liu2024improved} for multimodal tasks. Language tasks follow the TRACE order, with Py150 placed last to limit its impact on intrinsic capabilities; multimodal tasks are sequenced by their effect on original capabilities: action, commonsense, count. All models use the Adam optimizer (batch size 64, seed 42), with learning rates: Gemma2-2B, 5e-4 for LoRA/O-LoRA, and 5e-5 for the others; for the other models, 1e-4 for LoRA/O-LoRA, and 1e-5 for the rest. No warmup is applied. Qwen3-14B is trained 1 epoch per task, others 5 epochs. Experiments run on 4×80G A100, 4×80G A800, and 8×40G A100 GPUs. See supplementary material B for baseline-specific deployment details.
\subsection{Main Results} \label{sec:analysis}
The main results are shown in Table~\ref{tab1} and Table~\ref{tab2}. Table~\ref{tab1} presents language tasks performance, and Table~\ref{tab2} reports multimodal results. Based on these results, we can draw the following conclusions:\\
\textbf{PIECE demonstrates outstanding performance across both language and multimodal tasks.} Across models of different scales, especially for large models, both PIECE-F and PIECE-S achieve the highest OP, outperforming all baselines. Meanwhile, the BWT results indicate that PIECE substantially alleviates forgetting during sequential task learning, highlighting its strong potential to balance new task acquisition and previous knowledge retention.\\
\textbf{PIECE effectively preserves the original capabilities of foundation models.} On the HumanEval benchmark for language models and the Flickr30K benchmark for multimodal models, PIECE significantly outperforms traditional approaches, verifying its robustness in maintaining general capabilities. Even for Qwen3-14B with one-epoch-per-task training and naturally mild forgetting, PIECE still maintains a clear advantage, demonstrating its consistent ability-preserving effect under different training intensities. \\
\textbf{PIECE-F excels in task transfer, while PIECE-S achieves stronger forgetting mitigation.} As reflected by the OP metric, PIECE-F slightly outperforms PIECE-S across several tasks, showing greater adaptability and transferability to new domains. Conversely, PIECE-S consistently achieves the best BWT results with the least forgetting. This stability not only mitigates performance degradation in sequential learning but also further contributes to preserving the foundation model’s original abilities.
\subsection{Analysis}
To further understand the behavior of \textbf{PIECE} beyond its overall performance, we conduct a series of analytical experiments focusing on three key questions:
\begin{enumerate}
    \item \emph{Where PIECE learns and why it achieves its objective?}
    \item \emph{Whether PIECE forgets more as task count increases?}
    \item \emph{How the choice of sparsity ratio affects performance?}
\end{enumerate}
All analyses are performed on the \textbf{Llama3-8B} model.\\
\colorbox{gray!10}{\emph{Where PIECE learns and why it achieves its objective?}}
\begin{figure}[!t]
    \centering
    \includegraphics[width=0.9\linewidth]{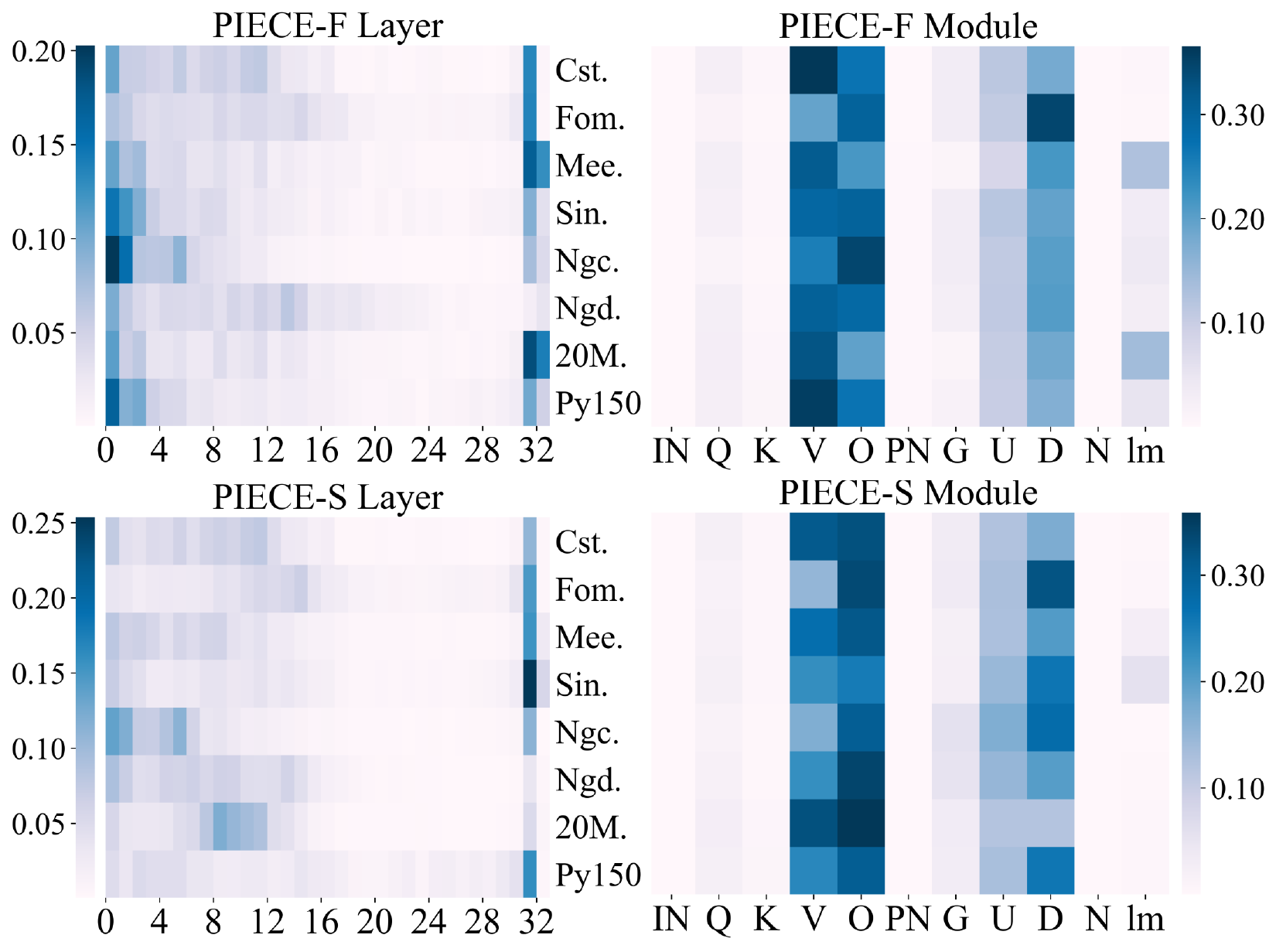}
    \caption{Distribution of critical parameters identified by \textbf{PIECE-F} and \textbf{PIECE-S} across different tasks.}
    \vspace{-4mm}
    \label{fig:result4_1}
\end{figure}
\begin{figure}[]
    \centering
    \includegraphics[width=0.9\linewidth]{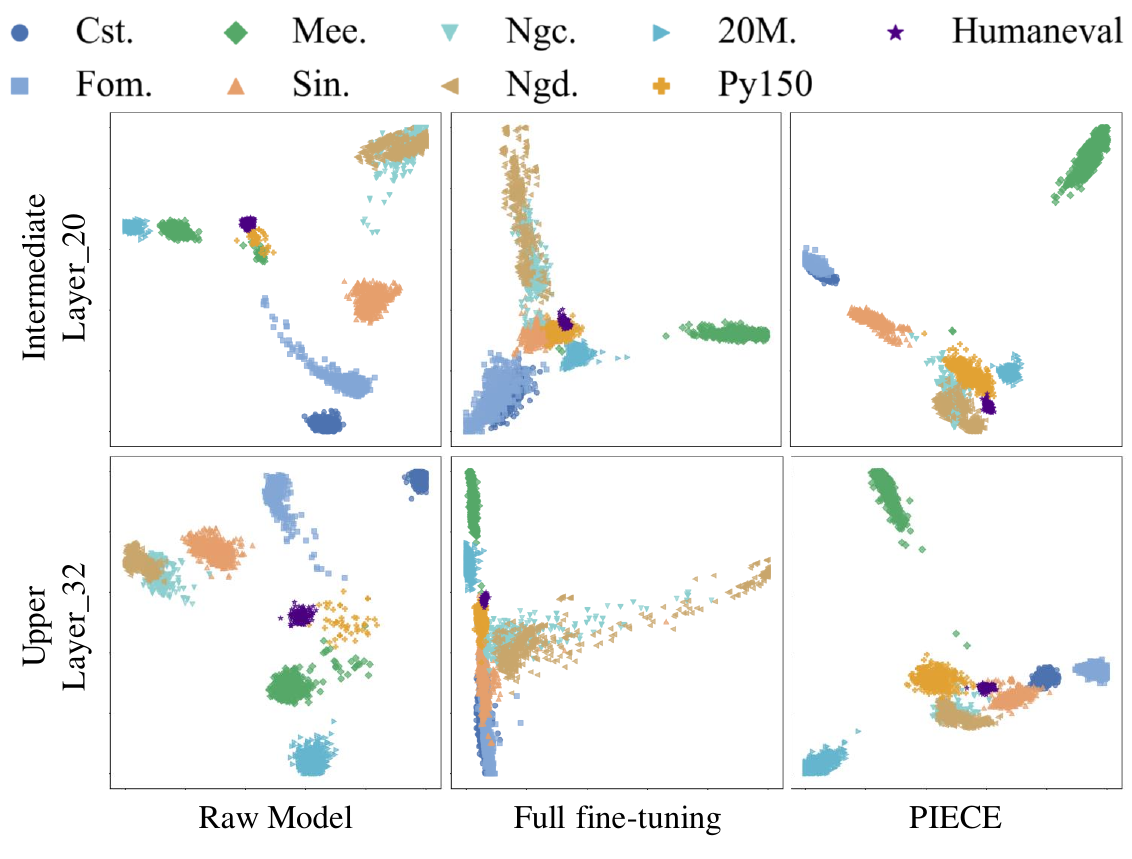}
    \caption{Visualization of intermediate and upper layer task representations comparing the base model, full fine-tuning, and \textbf{PIECE}.}
    \label{fig:result4_2}
    \vspace{-4mm}
\end{figure}

To explore where \textbf{PIECE} learns and why it is stable, we visualize the important parameters selected by \textbf{PIECE-F} and \textbf{PIECE-S} across tasks. Figure~\ref{fig:result4_1} shows both methods mainly target attention submodules $\texttt{V}/\texttt{O}$ and feed-forward layers ($\texttt{U}/\texttt{D}$). This aligns with \citet{yao2024theoretical}, which suggests that the $\texttt{V}/\texttt{O}$ modules play a more crucial role than $\texttt{Q}/\texttt{K}$ in downstream adaptation, and \citet{dai2022knowledge}, which identifies $\texttt{U}$ and $\texttt{D}$ as key locations for knowledge storage. At the layer level, both methods primarily attend to the higher layers, consistent with \citet{zhao2024layer} linking them to downstream task relevance. The distinction is that \textbf{PIECE-F} also emphasizes lower layers, potentially aiding better adaptation to downstream tasks, although adjusting these shared modules may introduce some forgetting. Notably, both mechanisms avoid the intermediate layers, suggesting that these layers encode core pretraining knowledge and are crucial for maintaining the structural stability of the model. To further examine the role of intermediate layers in the model’s multitask capabilities, we visualized task representations at intermediate and upper layers using dimensionality reduction. As shown in Figure~\ref{fig:result4_2}, compared to base model, full fine-tuning disrupts the representations in intermediate layers, consequently impairing the final task separability, whereas \textbf{PIECE} preserves clear task distinctions.\\
\colorbox{gray!10}{\emph{Whether PIECE forgets more as task count increases?}}
\begin{figure}[]
    \centering
    \includegraphics[width=\linewidth]{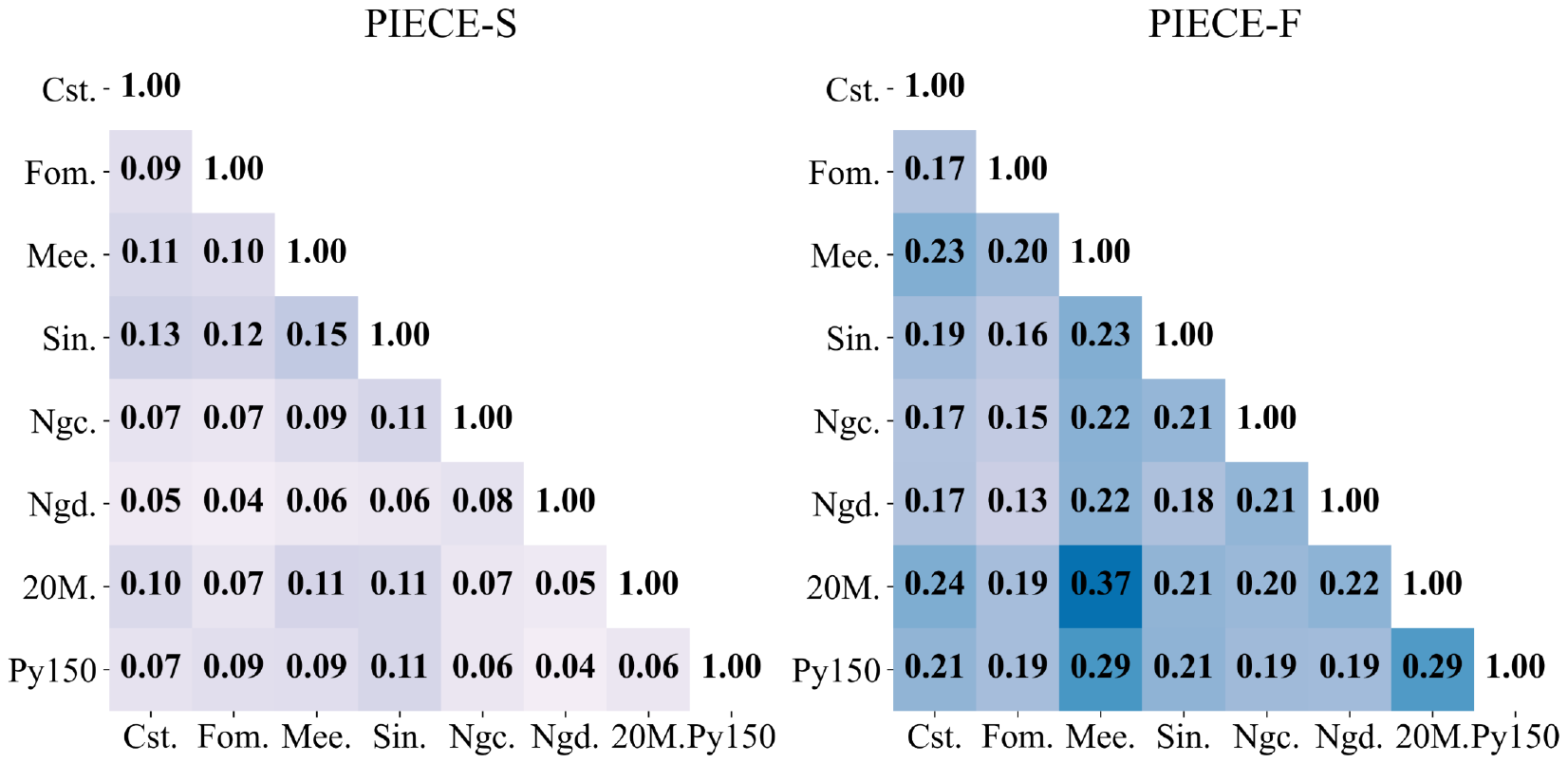}
    \caption{Parameter overlap across tasks.}
    \label{fig:result4_3}
    \vspace{-4mm}
\end{figure}
\begin{figure}[]
    \centering
    \includegraphics[width=0.9\linewidth]{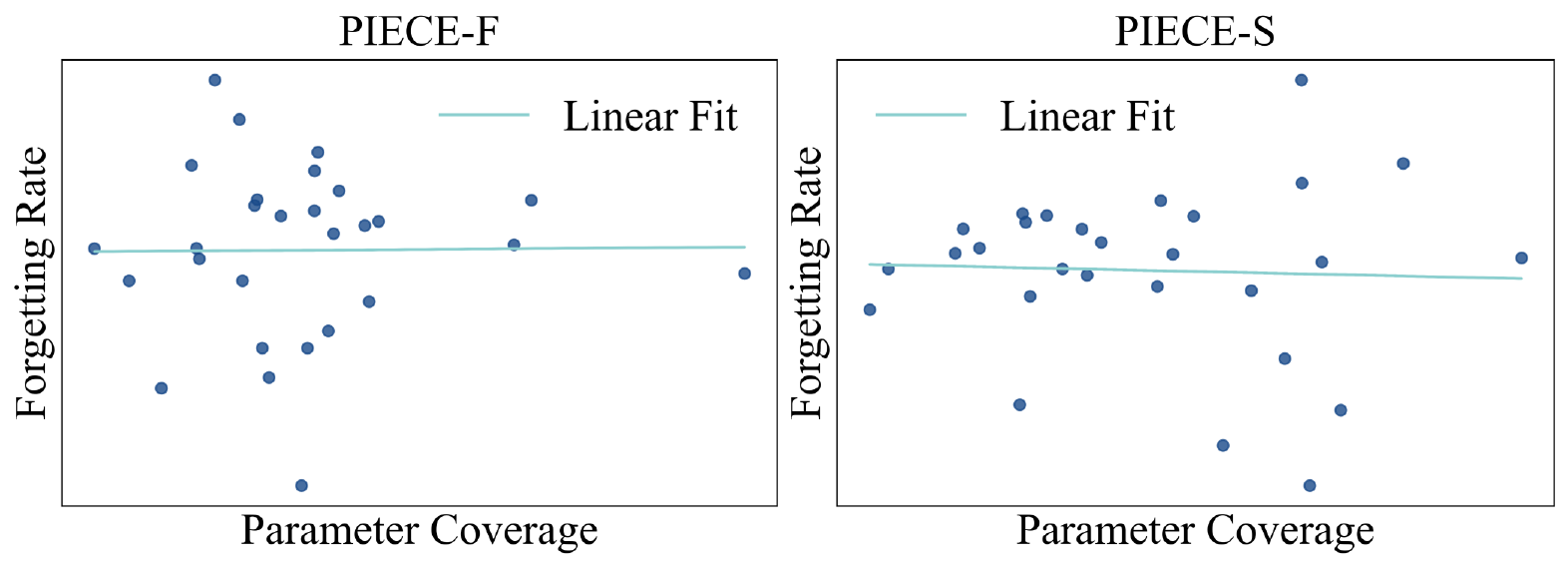}
    \caption{Correlation of parameter overlap and forgetting rate.}
    \label{fig:result4_4}
    \vspace{-4mm}
\end{figure}

The core of \textbf{PIECE} lies in fine-tuning only a small subset of critical parameters, thereby mitigating catastrophic forgetting. This raises a natural question: as the number of tasks increases, do important parameters overlap significantly across tasks, potentially increasing the risk of forgetting? To investigate the scalability of \textbf{PIECE}, we first analyze the parameter overlap across tasks. As shown in Figure~\ref{fig:result4_3}, the overlap is relatively low for both \textbf{PIECE-F} and \textbf{PIECE-S}, with \textbf{PIECE-S}, which emphasizes task-specific parameters, exhibiting the lowest overlap. Moreover, we compute the correlation between parameter overlap and task forgetting rate (Figure~\ref{fig:result4_4}). The results show negligible correlation for both methods (Pearson r: -0.04 for \textbf{PIECE-S}, 0.01 for \textbf{PIECE-F}), indicating that even as the number of tasks grows, \textbf{PIECE} can effectively preserve critical parameters and maintain performance on previous tasks. \\
\colorbox{gray!10}{\emph{How the choice of sparsity ratio affects performance?}}
\begin{figure}[]
    \centering
    \includegraphics[width=\linewidth]{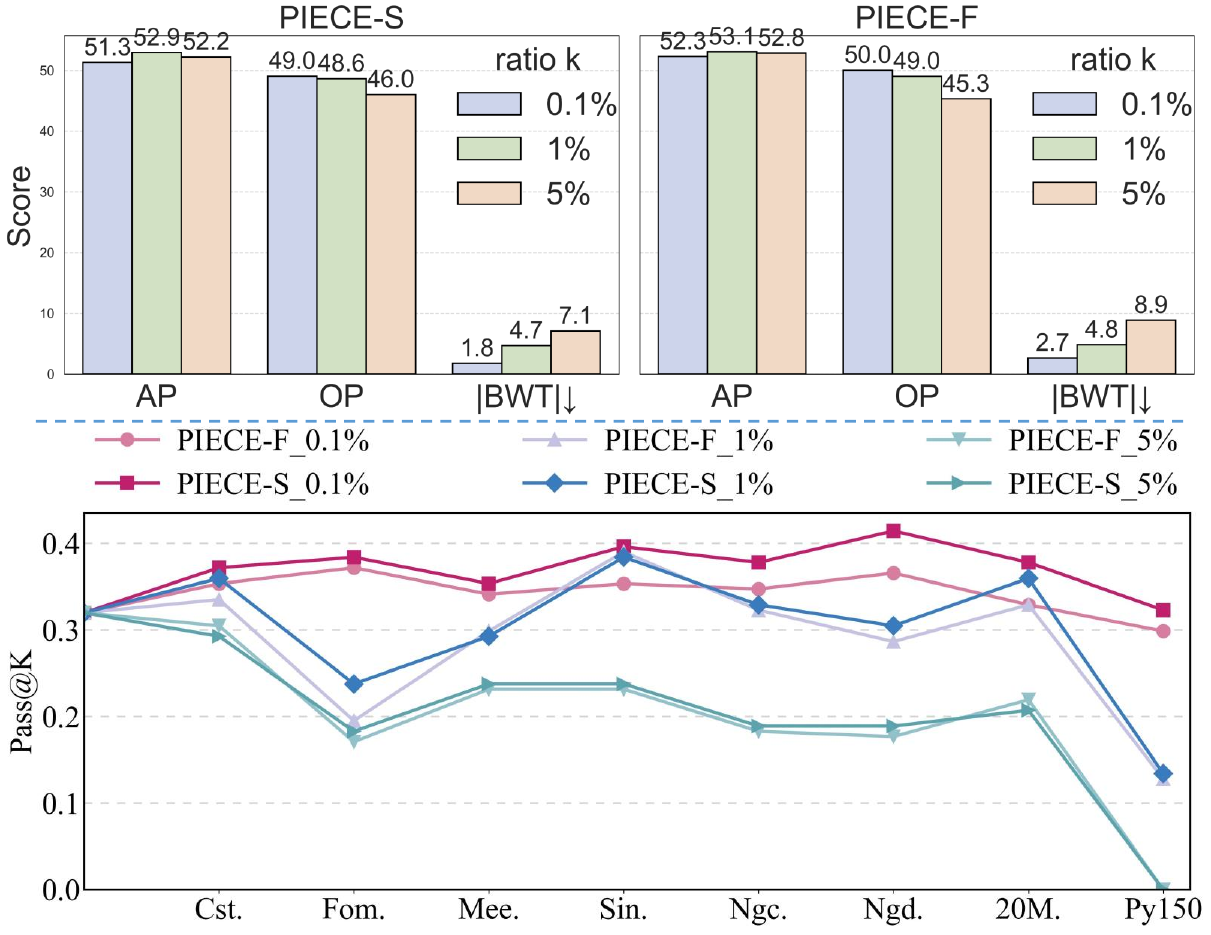}
    \caption{Effect of the number of updated parameters on performance and forgetting. Average Performance (AP) is the mean performance across sequential tasks after each task, reflecting transfer ability: \(AP_t = \frac{1}{t}\sum_{i=1}^t R_{i,i}\)}
    \label{fig:result4_5}
    \vspace{-4mm}
\end{figure}

To investigate the effect of the number of updated parameters on performance, we varied the selection ratio \(k\), with the results shown in Figure~\ref{fig:result4_5}. The experiments indicate that increasing \(k\) initially improves the model’s transfer performance on downstream tasks, but this improvement is not sustained; at \(k=5\%\), performance is actually lower than at \(k=1\%\). Meanwhile, larger \(k\) values substantially exacerbate forgetting, resulting in a decrease in OP and causing more severe degradation of the    original capabilities.
\section{Conclusion}\label{sec:conclusion}
We present PIECE, a Parameter Importance Estimation-based Continual Enhancement method for foundation models. PIECE mitigates catastrophic forgetting by selectively updating only the most critical 0.1\% of parameters, guided by theoretically grounded importance estimation. Without relying on replay data, task labels, or architectural modifications, PIECE achieves efficient domain adaptation while preserving general reasoning and programming abilities. Extensive experiments on diverse LLMs and MLLMs show that PIECE surpasses existing approaches in both retention and transfer. By coupling parameter-efficient tuning with importance estimation, PIECE offers a scalable, model-agnostic pathway toward sustainable, domain-adaptive foundation models. In the future, we will further distinguish fine-tuning key parameters from stability-critical ones to better balance downstream transfer and forgetting mitigation.

\section{Acknowledgments}\label{sec:ack}
This work was funded by the National Natural Science Foundation of China (NSFC) under Grants No. 62406013, the Beijing Advanced Innovation Center Funds for Future Blockchain and Privacy Computing(GJJ-24-034), and the Fundamental Research Funds for the Central Universities.
{
    \small
    \bibliographystyle{ieeenat_fullname}
    \bibliography{main}
}
\clearpage
\setcounter{page}{1}
\maketitlesupplementary


%

\section*{\textbf{A} Datasets}
\subsection*{A.1 Language Task Details}
\begin{table*}[!t]
\centering
\resizebox{\textwidth}{!}{
\setlength{\tabcolsep}{1.4mm}{
\begin{tabular}{lcccccccccccccc}
\hlineB{4}
\multirow{2}{*}{Dataset} & \multicolumn{9}{c}{Language Task} & \multicolumn{4}{c}{Multi-modal Task}\\ 
\cline{2-15} & Cst. & Fom. & Mee. & Sin. & Ngc. & Ngd. & 20M. & Py150. & HumanEval & Act. & Com. & Cou. & Flickr30k &\\ \hline
Train & \multicolumn{8}{c}{1000} & - & \multicolumn{3}{c}{1000} & - & \\
Test & 2000 & 496 & 692 & 2000 & 81 & 325 & 200 & 2000 & 164 & 1448 & 1227 & 2905 & 1000 & \\
\hline
\end{tabular}}}
\caption{Dataset Statistics for Language and Multi-modal Tasks.}
\label{tabA_1}
\vspace{-4mm}
\end{table*}
The language tasks in this paper are derived from the TRACE dataset, which was constructed following three principles:
\begin{enumerate}
    \item the datasets should be sufficiently novel so that large language models (LLMs) exhibit weak initial performance;
    \item the tasks should be challenging enough to comprehensively test reasoning and generalization capabilities;
    \item the tasks should cover a wide range of domains and types.
\end{enumerate}
Due to its well-designed and diverse task settings, TRACE has been adopted in several recent continual learning studies~\cite{tong2025analytic, he2024seekr, feng2025recurrent, du2024unlocking}.
The language tasks in TRACE can be categorized into four groups: \textbf{Domain-Specific Tasks}, \textbf{Multi-lingual Tasks}, \textbf{Code QA Tasks}, and \textbf{Mathematical Reasoning Tasks}. The statistics for each category are shown in Table~\ref{tabA_1}, and configurations are introduced below.\\
\textbf{Domain-Specific Tasks} \\
Domain-specific tasks evaluate a model’s knowledge acquisition and application within specialized domains. When relevant domain information is underrepresented in training corpora, model performance can degrade significantly. TRACE includes the following three datasets under this category:
\begin{itemize}
    \item \textbf{ScienceQA (Sin.)}~\cite{lu2022learn}: Collected from elementary and high school science curricula, covering topics across natural science, social science, and language science. This dataset requires multi-hop reasoning and scientific knowledge understanding. Only text-only samples are retained in this work.
    \item \textbf{FOMC (Fom.)}~\cite{shah2023trillion}: A financial-domain classification task aimed at identifying whether Federal Reserve policy statements are “hawkish” or “dovish.” The original corpus consists of three parts—meeting minutes, press conference transcripts, and speeches. A combined version of all three is used in this work.
    \item \textbf{MeetingBank (Mee.)}~\cite{hu2023meetingbank}: A newly introduced dataset for city council meeting summarization, requiring global comprehension and compression of lengthy meeting transcripts. As a relatively unexplored domain, it effectively evaluates models’ long-context understanding and abstraction abilities.
\end{itemize}
\textbf{Multi-lingual Tasks} \\
Cross-lingual capability serves as an important measure of a large language model’s generalization ability. However, due to vocabulary coverage and corpus imbalance, most models perform worse on non-English languages. TRACE includes two datasets for evaluating such multi-lingual abilities:
\begin{itemize}
    \item \textbf{C-STANCE (Cst.)}~\cite{zhao2023c}: A Chinese stance detection dataset collected from Sina Weibo. It includes two subtasks: target-based and domain-based stance detection. We adopt the target-based subtask, where targets in the test set do not appear in the training set, to evaluate zero-shot and cross-domain generalization.
    \item \textbf{20Minuten (20M.)}~\cite{gonzales2021new}: A German text simplification dataset derived from the Swiss news magazine 20 Minuten. Each instance consists of an original article and its simplified summary, used to evaluate model performance in German text generation.
\end{itemize}
\textbf{Code QA Tasks} \\
Code-related tasks are used to assess a model’s understanding and generation capabilities under structured, long-context conditions. TRACE employs the Py150~\cite{lu1codexglue} corpus to construct the code QA task. The corpus contains approximately 150,000 Python programs collected from GitHub repositories. The task requires predicting the subsequent content given code context, where label formats differ from those in standard code generation tasks, posing additional challenges to capability retention. Considering that \textbf{HumanEval} is used in this paper to measure the retention of code generation ability, this task plays a crucial role in assessing model performance. Thus, we place Py150 last in TRACE testing order to limit its impact on intrinsic capabilities.\\
\textbf{Mathematical Reasoning Tasks} \\
Mathematical reasoning tasks are designed to test models’ arithmetic and logical reasoning capabilities. TRACE selects the first two subtasks (Ngc., Ngd.) from the \textbf{NumGLUE} benchmark~\cite{mishra2022numglue}. Both subtasks require numerical computation and logical inference in natural language settings. These tasks impose higher demands on symbolic reasoning and are particularly useful for evaluating capability retention under abstract reasoning scenarios.

Given that programming ability is widely recognized as a representative prior knowledge that is particularly susceptible to forgetting during post-training~\cite{chen2025towards}, we adopt it as a measure of the model’s retention of original capabilities and quantify it using the \textbf{HumanEval} benchmark. HumanEval is a standardized evaluation dataset for Python function generation tasks, designed to assess the code generation capabilities of large language models~\cite{li2025structured, beger2025coconut, zhang2025p3}. The dataset contains 164 programming problems, each consisting of a function description (function docstring) and a function signature. The model is required to generate a complete, executable function implementation based on the natural language description. Unlike typical natural language generation tasks, HumanEval evaluation relies on unit tests rather than manual annotation. The dataset is not used for training, and is solely employed to assess the model’s retention of original capabilities after multi-task continual learning. Since programming tasks demand a high degree of logical consistency, semantic accuracy, and structured reasoning, and these abilities are particularly vulnerable to degradation during multi-task training, performance changes on HumanEval provide a sensitive and reliable indicator of knowledge retention and forgetting in continual learning scenarios.

\subsection*{A.2 Multi-modal Task Details}
The multi-modal tasks in this paper are derived from the VQA v2 dataset~\cite{goyal2017making}. To facilitate continual learning experiments, we follow the task grouping scheme of VQACL~\cite{zhang2023vqacl}, splitting the tasks by question type and randomly sampling 1,000 training examples per task while retaining the original test sets.
Based on performance improvements and capability retention, we focus on the following three question types in this study:
\begin{itemize}
    \item \textbf{Action (Act.)}: Action recognition questions, which require the model to understand the actions of people or objects in an image;
    \item \textbf{Commonsense (Com.)}: Commonsense reasoning questions, which require the model to make judgments based on both visual information and general knowledge;
    \item \textbf{Count (Cou.)}: Counting questions, which require the model to accurately identify the number of target objects in an image.
\end{itemize}
In addition, we use the Flickr30K dataset~\cite{young2014image} to evaluate the model’s overall capability retention in multi-modal tasks. This dataset contains rich image–text pairs, with captions typically longer than those in VQA tasks, making it suitable for assessing the model’s comprehensive performance in multi-modal understanding and generation.

\section*{\textbf{B} Baselines and Implementation Details}
This section provides detailed descriptions of all continual learning baselines compared in this work, including their core ideas and optimization objectives. The baselines fall into three categories: \textbf{Regularization-based methods}, \textbf{Replay-based methods}, and \textbf{Parameter-Efficient Tuning (PET)}.
\subsection*{\textbf{B.1} Regularization-based Methods}
\textbf{EWC (Elastic Weight Consolidation).}\\
EWC~\cite{kirkpatrick2017overcoming} estimates the importance of model parameters on previous tasks and applies a quadratic penalty to discourage significant updates on important parameters when learning new tasks. The overall objective for task \(t\) is :
\begin{equation}\mathcal{L}_{\mathrm{EWC}}=\mathcal{L}_t(\theta)+\lambda\sum_iF_i(\theta_i-\theta_i^*)^2,\end{equation}
where \(\theta^*\) denotes the parameter values after training the previous task, \(F_i\) is the Fisher information representing the importance of parameter \(i\) and \(\lambda\) controls the strength of the regularizer (set to 0.5 in our experiments).\\
\textbf{GEM (Gradient Episodic Memory).}\\
GEM~\cite{lopez2017gradient} stores a small number of samples from previous tasks in an episodic memory and constrains gradient updates so that the loss on old tasks does not increase. During training on task \(t\), GEM requires:
\begin{equation}\langle g,g_k\rangle\geq0,\quad\forall k<t,\end{equation}
where \(g\) is the gradient for the current task and \(g_k\) is the gradient computed on memory samples of task \(k\). If this constraint is violated, i.e., \(\langle g,g_k\rangle<0\), GEM solves the following quadratic program to obtain a projected gradient:
\begin{equation}\begin{aligned}
 & \min_{\tilde{g}} & & \|\tilde{g}-g\|^{2} \\
 & \mathrm{s.t.} & & \langle\tilde{g},g_{k}\rangle\geq0,\forall k<t.
\end{aligned}\end{equation}
The projected gradient \(\tilde{g}\) is then used for parameter updates, ensuring no negative backward transfer.\\
\textbf{LwF (Learning without Forgetting).}\\
LwF~\cite{li2017learning} employs the predictions of the previous model as soft targets and introduces a distillation loss to constrain changes in the output distribution when learning new tasks:
\begin{equation}\mathcal{L}_{\mathrm{LwF}}=\mathcal{L}_t^{\mathrm{CE}}+\alpha\cdot\mathcal{L}_{\mathrm{KD}},\end{equation}
where the distillation loss is:
\begin{equation}\mathcal{L}_{\mathrm{KD}}=-\sum_cp_c^{\mathrm{old}}(T)\log p_c^{\mathrm{new}}(T),\end{equation}
with softened probabilities computed using temperature \(T\):
\begin{equation}p_c(T)=\frac{\exp(z_c/T)}{\sum_j\exp(z_j/T)}.\end{equation}
Following common practice, we set \(\alpha = 0.5\) and \(T=2\).

\subsection*{\textbf{B.2} Replay-based Methods}
Replay-based methods mitigate forgetting by reusing data from previous tasks when training new ones. We implement two representative replay strategies:
\textbf{Replay (offline replay).}
After completing training on the current task, we perform an additional replay phase where we uniformly sample 1\% of data from each previous task stored in a unified memory buffer. These real samples are used for extra gradient steps to reinforce previously learned knowledge.\\
\textbf{Replay-online (online replay)}
During each training iteration of a new task, we dynamically sample approximately 1\% of data from the same memory buffer and mix it with the current task’s mini-batch. This enables continuous rehearsal throughout training, improving stability in knowledge retention.

Both strategies rely solely on real data and share the same unified memory buffer. No synthetic data or auxiliary generative models are used. We also enforce uniform sampling across tasks to ensure fairness in the replay process.
\subsection*{\textbf{B.3} Parameter-Efficient Tuning (PET) Baselines}
\textbf{SeqLoRA (Sequential Low-Rank Adaptation)}:\\
LoRA~\cite{hulora} introduces a trainable low-rank update to the weight matrix of attention layers. For a linear transformation \(h=Wx\), LoRA reparameterizes the weight as:
\begin{equation}W^{\prime}=W+\Delta W,\quad\Delta W=BA,\end{equation}
where \(A\in\mathbb{R}^{r\times d},B\in\mathbb{R}^{d\times r},\), and \(r=8\) is the rank used in our experiments. A scaling factor \(\alpha\) is applied to stabilize optimization (set to 32): 
\begin{equation}\Delta W=\frac{\alpha}{r}BA,\end{equation}
We follow common practice and apply LoRA to the attention projection matrices q\_proj and v\_proj, with dropout set to 0.1.
SeqLoRA shares the same LoRA parameters across all tasks, enabling a simple sequential PET baseline.\\
\textbf{O-LoRA (Orthogonal LoRA).}\\
O-LoRA~\cite{wang2023orthogonal} prevents interference across tasks by enforcing orthogonality between the LoRA subspaces of different tasks. For LoRA parameters \(A_i\) and \(A_j\) of task \(i\) and \(j\), the orthogonality regularizer is:
\begin{equation}\mathcal{L}_{\mathrm{ortho}}=\gamma\left\|A_i^\top A_j\right\|_F^2.\end{equation}
and the final test-time weight combines LoRA updates from all tasks:
\begin{equation}W_{\mathrm{test}}=W_0+\sum_tB_tA_t.\end{equation}
LoRA hyperparameters follow the same settings as SeqLoRA, with orthogonality coefficient \(\gamma=0.5\).\\
\textbf{LayerNorm.} \\
LayerNorm~\cite{zhao2024tuning} updates only the LayerNorm parameters while keeping all other model weights frozen. \\
\textbf{MIGU (MagnItude-based Gradient Updating  for continual learning).} \\
MIGU~\cite{du2024unlocking} is an unstructured PET method that selectively updates parameters based on their output magnitude. For a linear layer with weight \({W}\in\mathbb{R}^{d_\mathrm{out}\times d_\mathrm{in}}\) and input \(x\), the response of the \(i\)-th output channel is:
\begin{equation}h_i={W}_i\cdot{x}\end{equation}
and its importance score is the L1 norm:
\begin{equation}n_i=\|h_i\|_1.\end{equation}
During training, only the top 0.1\% parameters ranked by these scores are updated, while gradients on the remaining parameters are masked out, making the update pattern consistent with our proposed method.

\section*{\textbf{C} Metrics}
Let the performance score of the model on task \(i\) after learning task \(t\) be denoted as \(R_{t,i}\). For classification tasks or question-answering tasks with single-token labels, the performance score is measured by \textbf{accuracy}; for other tasks, the performance score is the average of the \textbf{Rouge-L} and \textbf{BLEU} scores. We use the following metrics to assess the learning effects of different methods on sequential tasks:
\subsection*{\textbf{C.1 Overall Performance (OP)}} 
Overall performance measures the average performance of the model across all learned tasks after training task \(t\):
\begin{equation}OP_t=\frac{1}{t}\sum_{i=1}^tR_{t,i}\end{equation}
\subsection*{\textbf{C.2 Backward Transfer (BWT)}} 
Backward transfer quantifies how learning new tasks affects the performance on previously learned tasks:
\begin{equation}BWT_t=\frac{1}{t}\sum_{i=1}^{t-1}\left(R_{t,i}-R_{i,i}\right)\end{equation}
where \(R_{i,i}\) is the performance on task \(i\) immediately after its training. Positive BWT indicates that new task learning improves old tasks, while negative BWT indicates forgetting. 
\textbf{Note}: OP and BWT are calculated only on sequential tasks to enable fair comparison with other baselines.
\subsection*{\textbf{C.3 Original Capability Retention}} 
\textbf{Language Models (Programming Ability)}\\
For language models, we use the HumanEval dataset to evaluate code generation ability. Pass@K is defined as:
\begin{equation}\mathrm{Pass@K}=\frac{1}{N}\sum_{i=1}^{N}\mathbb{I}(y_{i}\in\{\hat{y}_{i}^{(1)},\hat{y}_{i}^{(2)},\ldots,\hat{y}_{i}^{(K)}\}), \end{equation}
where \(y_i\) is the ground-truth code for the \(i\)-th test case, \(\hat{y}_i^{(k)}\) is the \(k\)-th generated sample, and \(K=1\) in this study.\\
\textbf{Multimodal Models (Image Captioning Ability)}\\
For multimodal models, we use the Flickr30K dataset, with performance measured as the average of Rouge-L and BLEU scores. BLEU-N is definded as:
\begin{equation}
\begin{aligned}
\mathrm{BLEU}_{N} &= \mathrm{BP} \cdot \exp\left(\frac{1}{N} \sum_{n=1}^{N} \log p_n \right),\\
\mathrm{BP} &=
\begin{cases}
1, & c > r \\
e^{1-r/c}, & c \le r
\end{cases}
\end{aligned}
\end{equation}
where \(c\) is the length of the generated sequence, \(r\) is the reference sequence length, and \(p_n\) is the n-gram precision.
Rouge-L is definded as:
\begin{equation}
\begin{aligned}
\mathrm{ROUGE\text{-}L} &= \frac{(1+\beta^2) \cdot P \cdot R}{R + \beta^2 \cdot P},\\
P &= \frac{\mathrm{LCS}(X,Y)}{|X|},\\
R &= \frac{\mathrm{LCS}(X,Y)}{|Y|},\\
\beta &= 1
\end{aligned}
\end{equation}
where \(X\) is the generated sequence, \(Y\) is the reference sequence, and \(\operatorname{LCS}(X,Y)\) denotes the length of the longest common subsequence.

\section*{\textbf{D} Extended Experimental Analysis}
This section provides additional experimental details and analyses, including the distribution of key parameters in the PIECE-F and PIECE-S models, as well as the overlap of importance parameters across tasks and their corresponding correlation coefficients, and it presents the PIECE training logs of selected models, to help readers gain a more comprehensive understanding of how the proposed method operates.

\begin{figure}[!t]
    \centering
    \includegraphics[width=\linewidth]{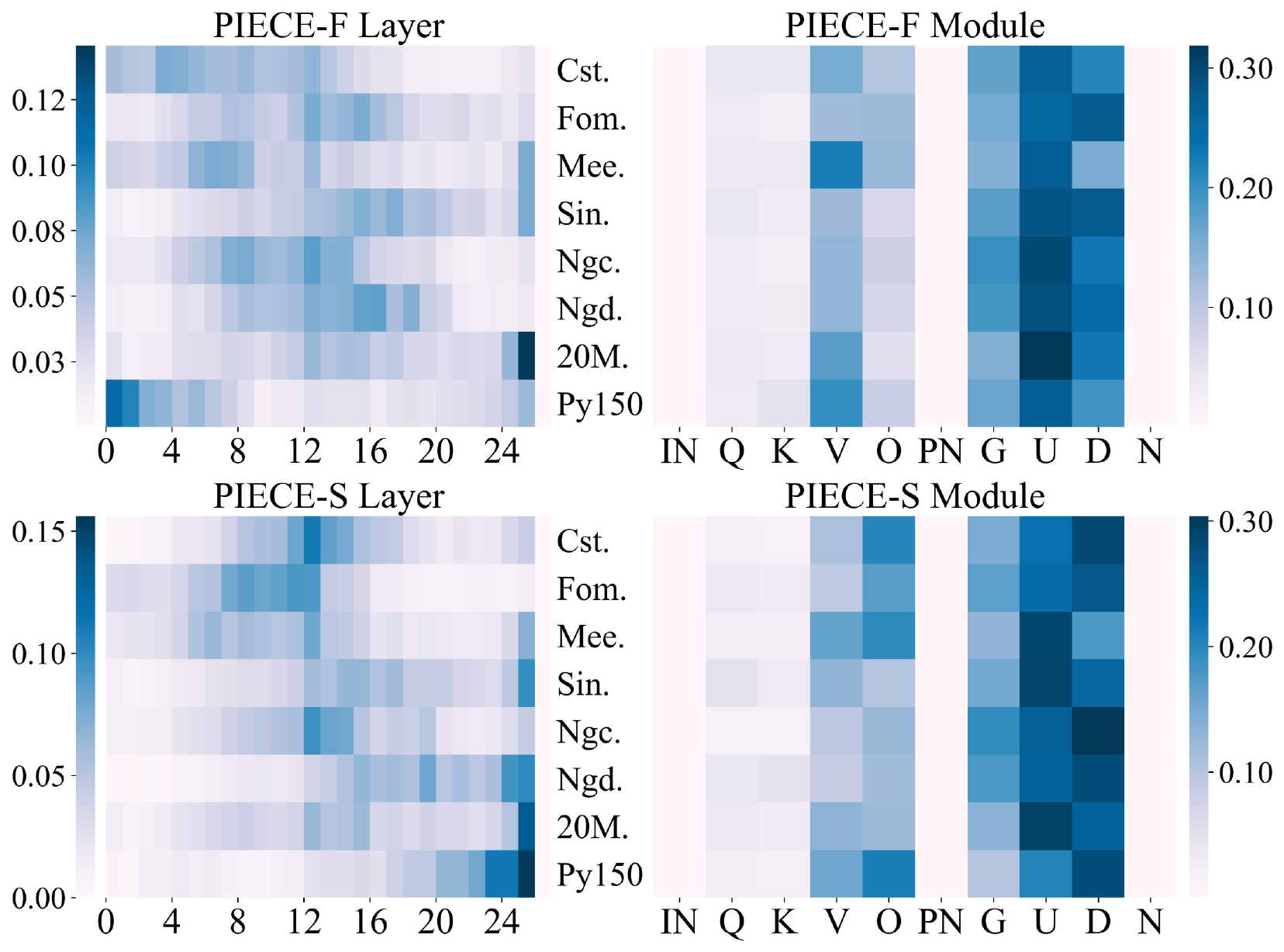}
    \caption{Distribution of critical parameters under Gemma2-2B.}
    \label{fig:result_1_2b}
\end{figure}

\begin{figure}[!t]
    \centering
    \includegraphics[width=\linewidth]{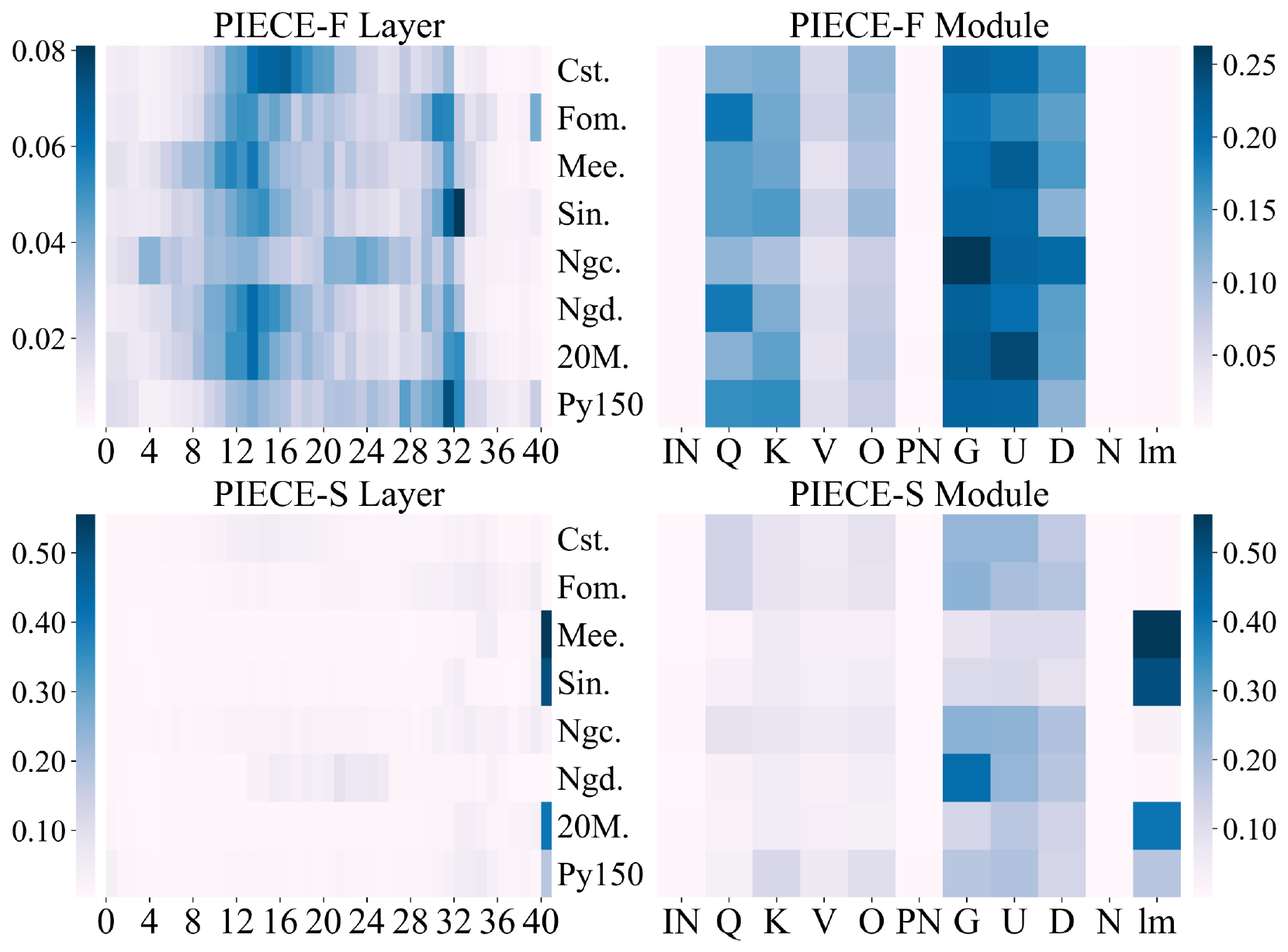}
    \caption{Distribution of critical parameters under Qwen3-14B (\(k=0.1\%\)).}
    \label{fig:result_1_14b_001}
\end{figure}

\subsection*{\textbf{D.1} Distribution of Critical Parameters}
\begin{figure}[!t]
    \centering
    \includegraphics[width=\linewidth]{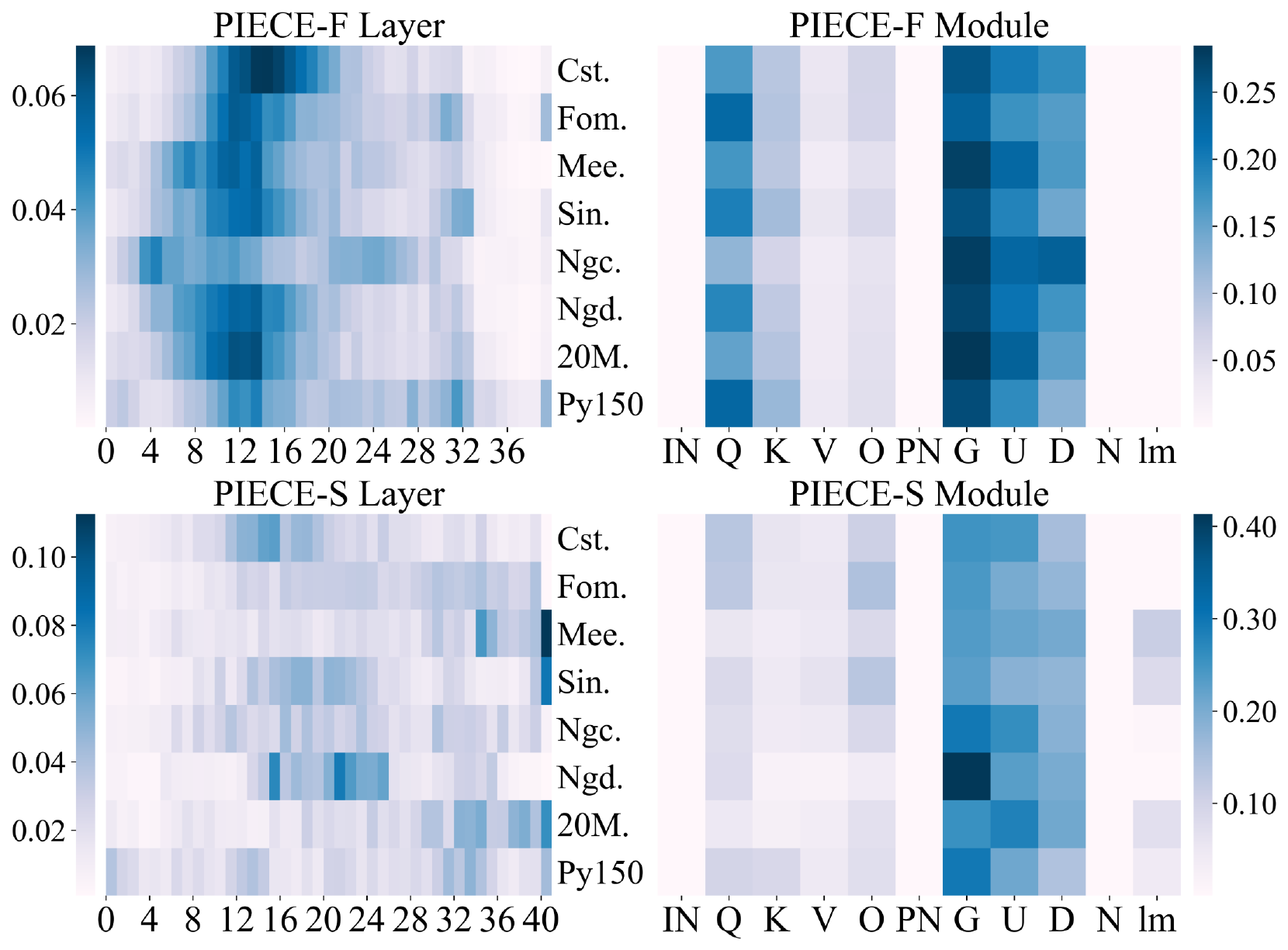}
    \caption{Distribution of critical parameters under Qwen3-14B (\(k=1\%\)).}
    \label{fig:result_1_14b}
\end{figure}

\begin{figure}[]
    \centering
    \includegraphics[width=\linewidth]{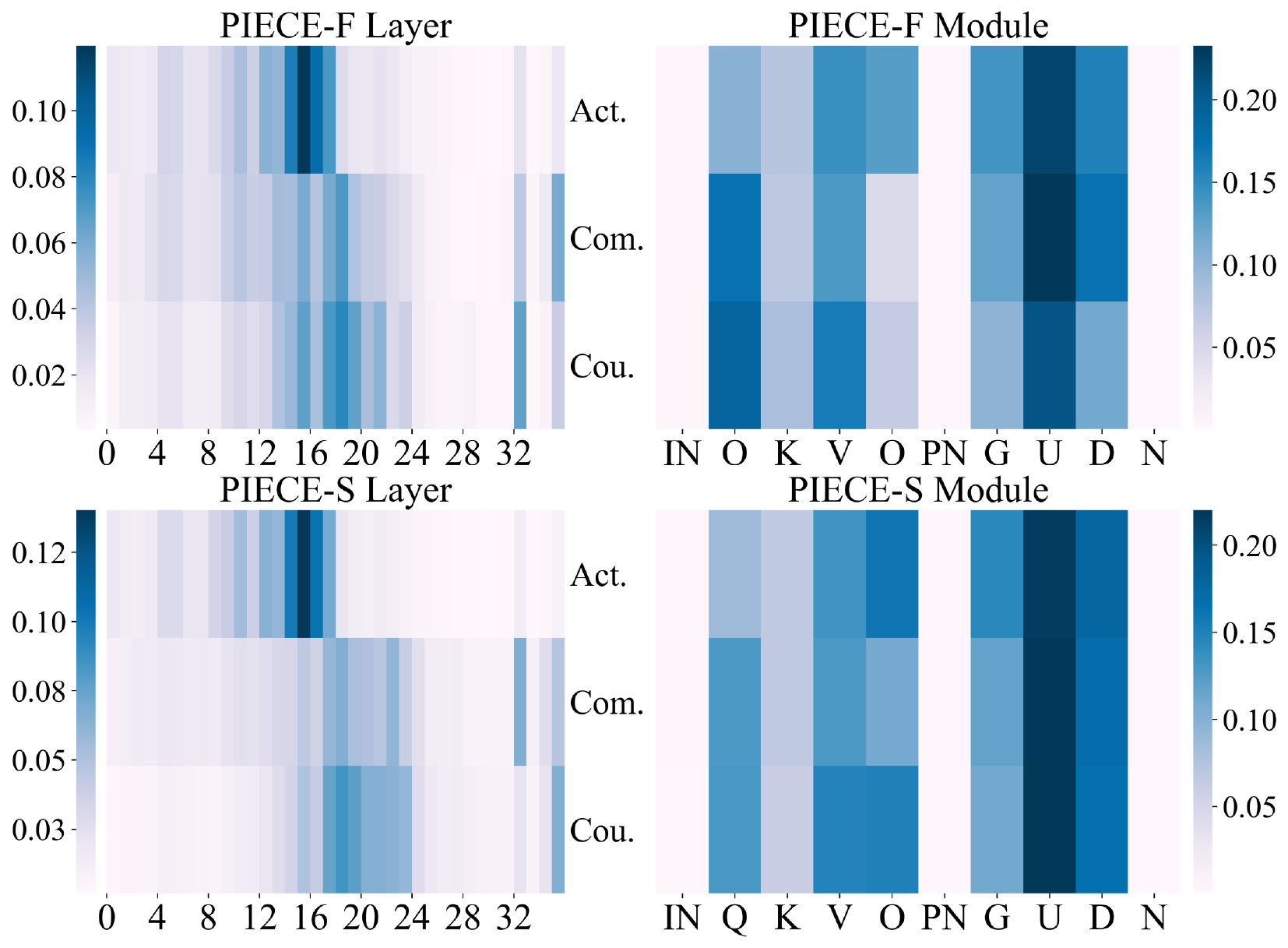}
    \caption{Distribution of critical parameters under Qwen3-VL-4B.}
    \label{fig:result_1_4b}
\end{figure}

\begin{figure}[]
    \centering
    \includegraphics[width=\linewidth]{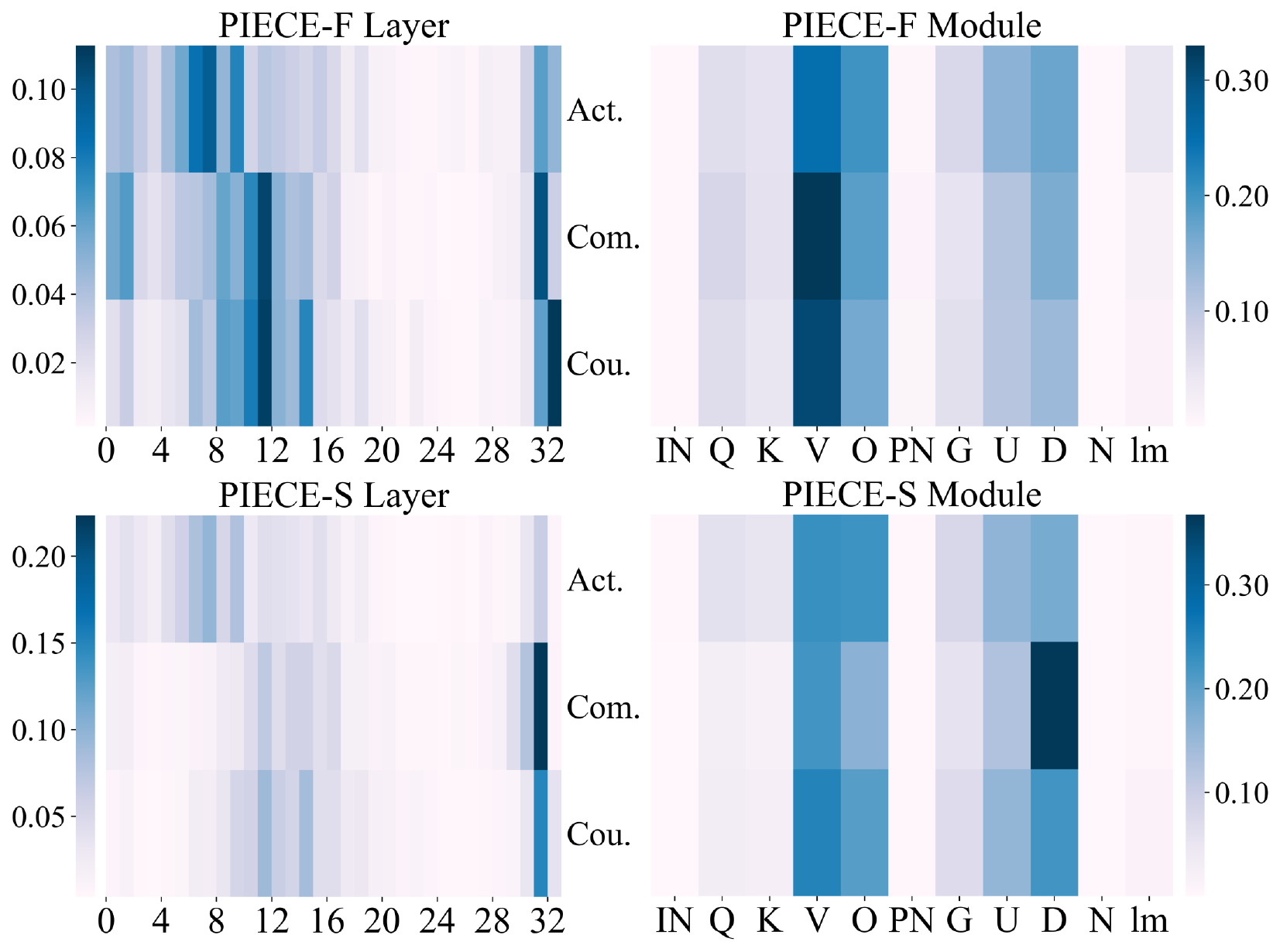}
    \caption{Distribution of critical parameters under LLaVA-1.5-7B.}
    \label{fig:result_1_7b}
\end{figure}

Figure~\ref{fig:result_1_2b}-\ref{fig:result_1_7b} illustrate the distribution of critical parameters across different models. 
We adopt a unified set of abbreviations for the major network modules within the model: the attention-related modules (q\_proj, k\_proj, v\_proj, o\_proj) are denoted as Q, K, V, and O, respectively; the feed-forward modules gate\_proj, up\_proj, and down\_proj are denoted as G, U, and D; the normalization layers input\_layernorm, post\_attention\_layernorm, and norm are denoted as IN, PN, and N; and the final output layer lm\_head is denoted as lm. These abbreviations are used consistently throughout the paper. To better observe the important parameters in large scale Qwen3-14B model, we expanded the proportion of selected important parameters from 0.1\% to 1\%, as shown in Figure~\ref{fig:result_1_14b_001} and \ref{fig:result_1_14b}.

Experimental results indicate that \textbf{architectural characteristics} and \textbf{task attributes} all potentially influence the distribution of important parameters within a network.
First, \textbf{from the perspective of model architecture}, the distribution of important parameters differs significantly across architectures, while models within the same architecture exhibit a high degree of similarity. For example, despite notable differences in task types, the distributions of important parameters across layers and submodules in Qwen3-VL-4B and Qwen3-14B are largely consistent in Figure~\ref{fig:result_1_14b} and \ref{fig:result_1_4b}. Similarly, LLaVA-1.5-7B and Llama3-8B, which share the LLaMA architecture, show comparable patterns in Figure~\ref{fig:result4_1} and \ref{fig:result_1_7b}. Moreover, Gemma2-2B has the similar updating submodules with LLaMA architecture in Figure~\ref{fig:result_1_2b}.

Second, \textbf{tasks attributes likely constitute another key factor influencing the distribution of important parameters.} Take the Qwen series as an example, although the 4B VLMs and 14B LMs share the same architecture, their preferences within feed-forward modules differ in Figure~\ref{fig:result_1_14b} and \ref{fig:result_1_4b}. The 4B VLMs tends to select higher-importance parameters in the up-proj (U) module, whereas the 14B LMs shows greater involvement in the gate-proj (G) module.


Although different models may vary in their objectives and intensity of parameter “protection” due to architectural and task differences, \textbf{all models consistently avoid updating certain parameters across layers and modules.} We posit that this parameter protection mechanism is a key reason why unstructured parameter-efficient fine-tuning achieves strong performance in continual learning.

Taken together, these observations suggest that model architecture and task attributes jointly influence the distribution of important parameters. While these hypotheses require further systematic investigation, they at least indicate that \textbf{evaluating important parameters through manual design or fixed heuristics is extremely challenging}, and they further highlight the practical significance of PIECE.


\subsection*{\textbf{D.2} Distribution Overlap and Correlation Across Tasks}
\begin{figure}[!t]
    \centering
    \includegraphics[width=\linewidth]{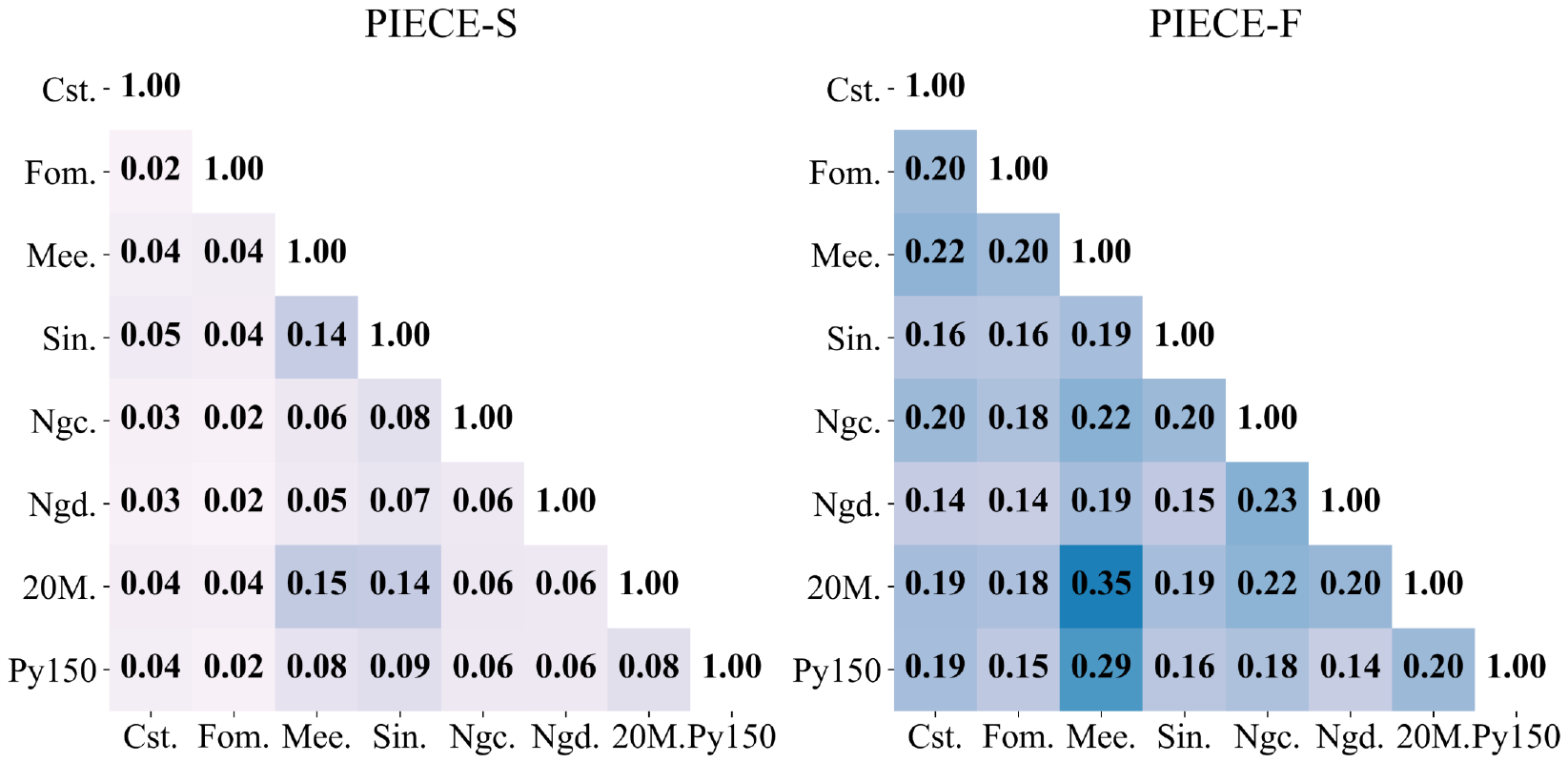}
    \caption{Parameter overlap across tasks under Gemma2-2B.}
    \label{fig:result_2_2b}
\end{figure}

\begin{figure}[!t]
    \centering
    \includegraphics[width=\linewidth]{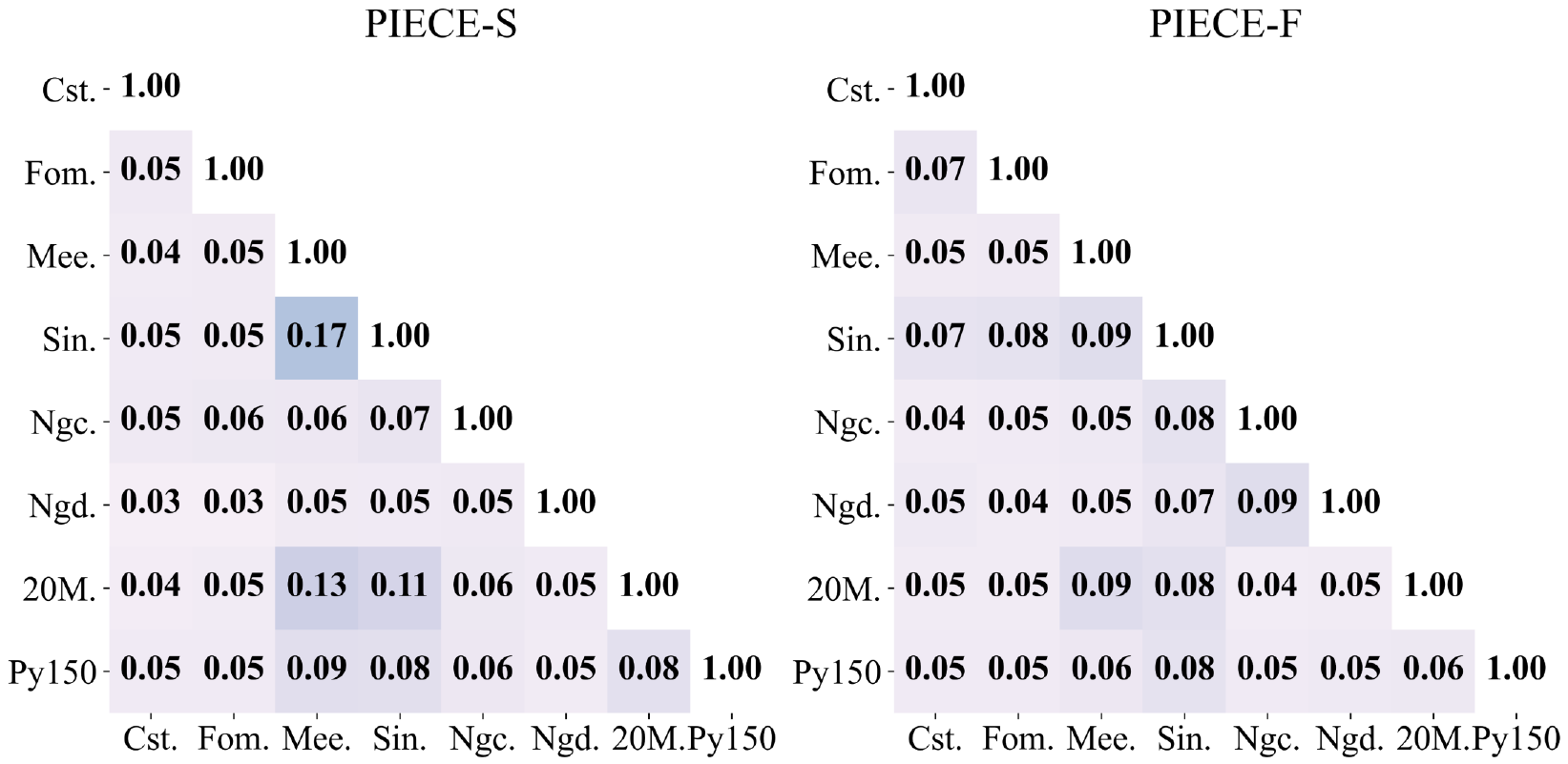}
    \caption{Parameter overlap across tasks under Qwen3-14B.}
    \label{fig:result_2_14b}
\end{figure}

\begin{figure}[!t]
    \centering
    \includegraphics[width=\linewidth]{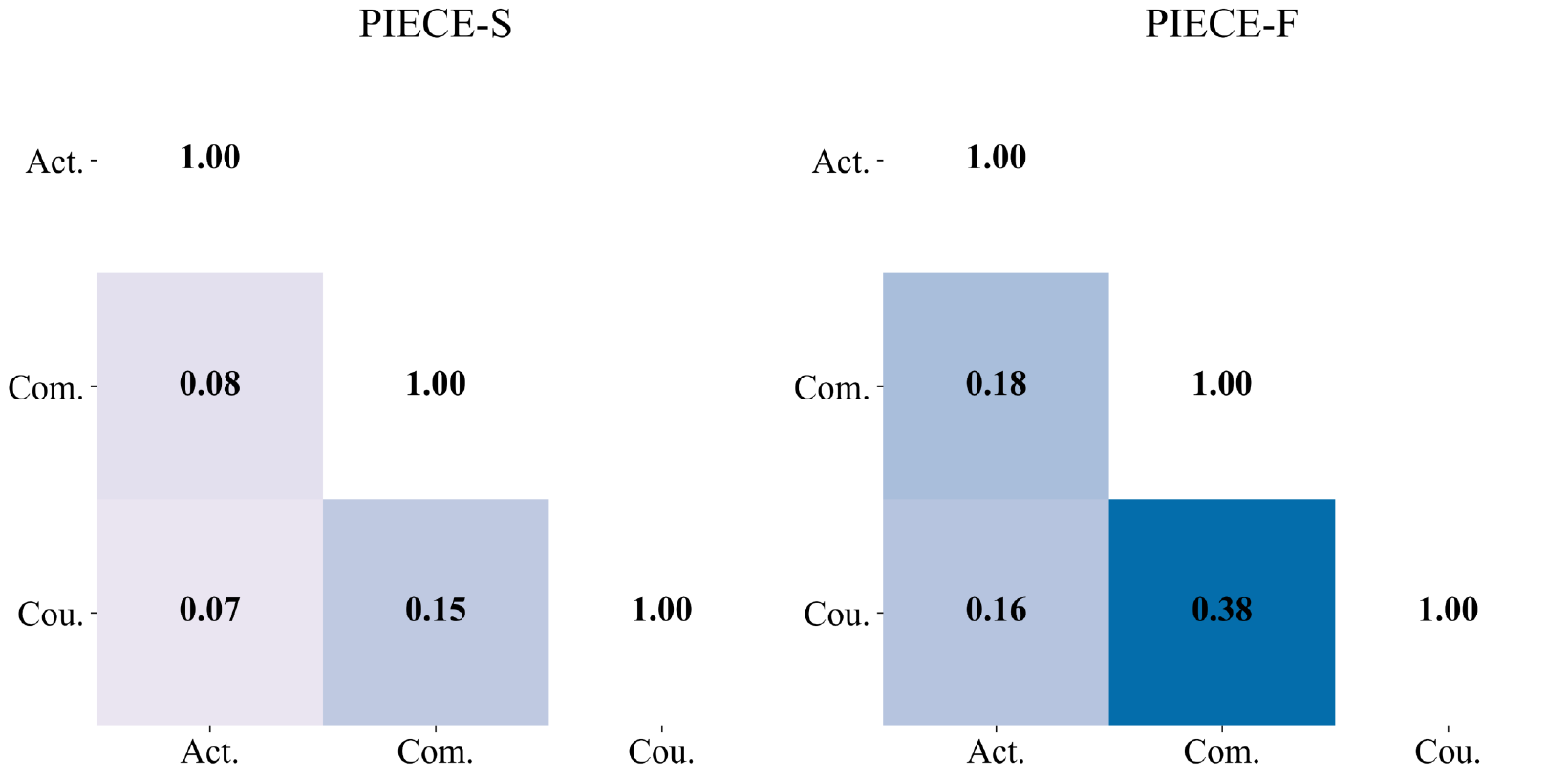}
    \caption{Parameter overlap across tasks under Qwen3-VL-4B.}
    \label{fig:result_2_4b}
\end{figure}

\begin{figure}[!t]
    \centering
    \includegraphics[width=\linewidth]{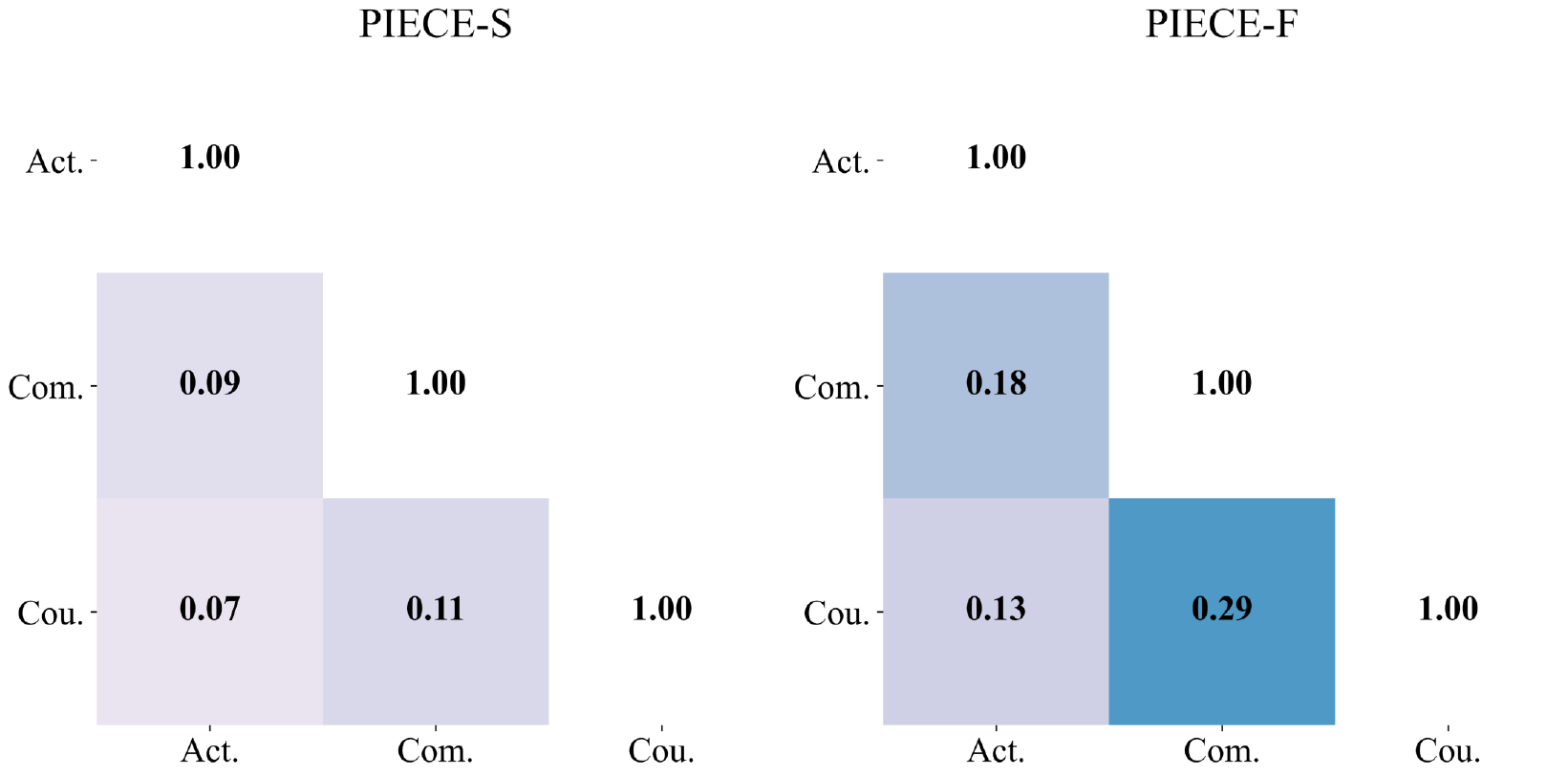}
    \caption{Parameter overlap across tasks under LLaVA-1.5-7B.}
    \label{fig:result_2_7b}
\end{figure}

\begin{table}[!t]
\centering
\resizebox{\linewidth}{!}{
\setlength{\tabcolsep}{1.4mm}{
\begin{tabular}{lcccccc}
\hlineB{4}
\multirow{2}{*}{} & \multicolumn{2}{c}{Gemma2-2B} & \multicolumn{2}{c}{Llama3-8B} & \multicolumn{2}{c}{Qwen3-14B}\\ 
\cline{2-7} & PIECE-F & PIECE-S & PIECE-F & PIECE-S & PIECE-F & PIECE-S\\ \hline
Pearson \(r\) & 0.03 & -0.06 & 0.01 & -0.04 & -0.39 & -0.30 \\
Pearson p & 0.88 & 0.75 & 0.96 & 0.83 & 0.45 & 0.39 \\
Spearman \(\rho\) & 0.01 & -0.17 & 0.06 & -0.01 & -0.42 & -0.29\\
Spearman p & 0.97 & 0.39 & 0.77 & 0.98 & 0.39 & 0.40\\
\hline
\end{tabular}}}
\caption{Importance Parameter Correlations Across Tasks. Pearson \(r\) and Spearman \(\rho\) indicate correlation strength (closer to ±1 = stronger correlation, near 0 = weak/no correlation); p-values indicate significance (closer to 0 = significant, near 1 = not significant).}
\label{tabD_1}
\end{table}
Figures~\ref{fig:result_2_2b}-\ref{fig:result_2_7b} illustrate the coverage of important parameters across tasks for various models. The results indicate that, in general, as model size increases, the proportion of parameters shared across tasks decreases, which aligns with intuition. Nevertheless, even for the smaller 2B model (Figure~\ref{fig:result_2_2b}), task-wise coverage remains relatively low. Across all model sizes, PIECE-S consistently exhibits lower coverage than PIECE-F, indicating greater task specificity in parameter allocation.

Moreover, although the absolute coverage values differ across models, the relative overlap patterns between tasks remain largely consistent. For instance, if two tasks show higher parameter overlap than other task pairs in one model, this relative trend tends to persist in other models as well, suggesting that certain task relationships are reflected robustly in parameter sharing patterns, independent of model size.

Table~\ref{tabD_1} further quantifies the relationship between coverage and task forgetting, showing that task forgetting is weakly or very weakly correlated with the extent of parameter overlap. This indicates that while overlap exists between important parameters across tasks, it does not directly predict forgetting, highlighting the complex relationship between parameter sharing and continual learning performance.

\subsection*{\textbf{D.3} Training Logs and Parameter Dynamics}
\begin{figure*}[!t]
    \centering
    \includegraphics[width=\textwidth]{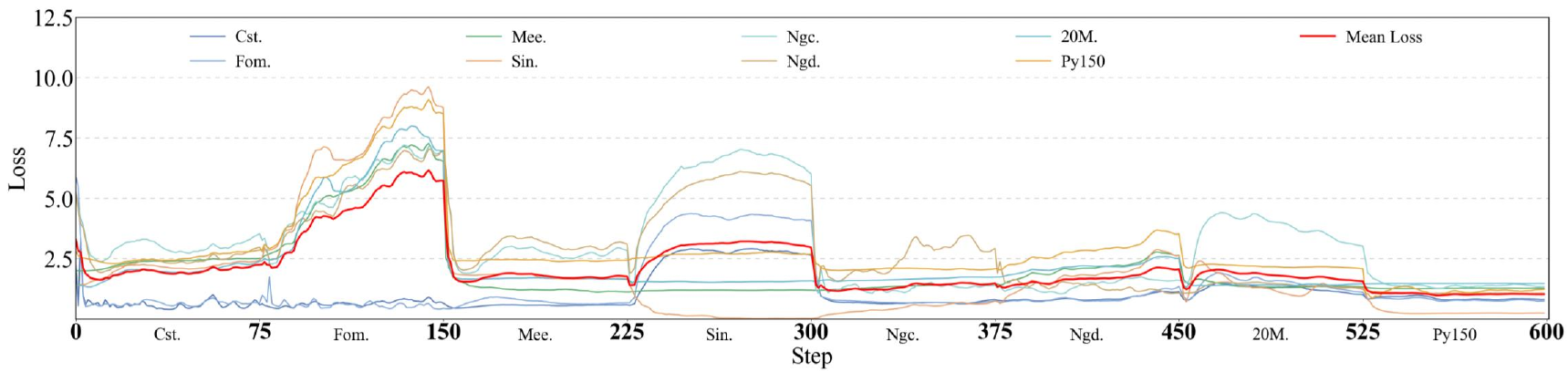}
    \caption{Multi-task Training Loss Curves for Llama3-8B (SeqFT).}
    \label{fig:llama3_full}
\end{figure*}

\begin{figure*}[!t]
    \centering
    \includegraphics[width=\textwidth]{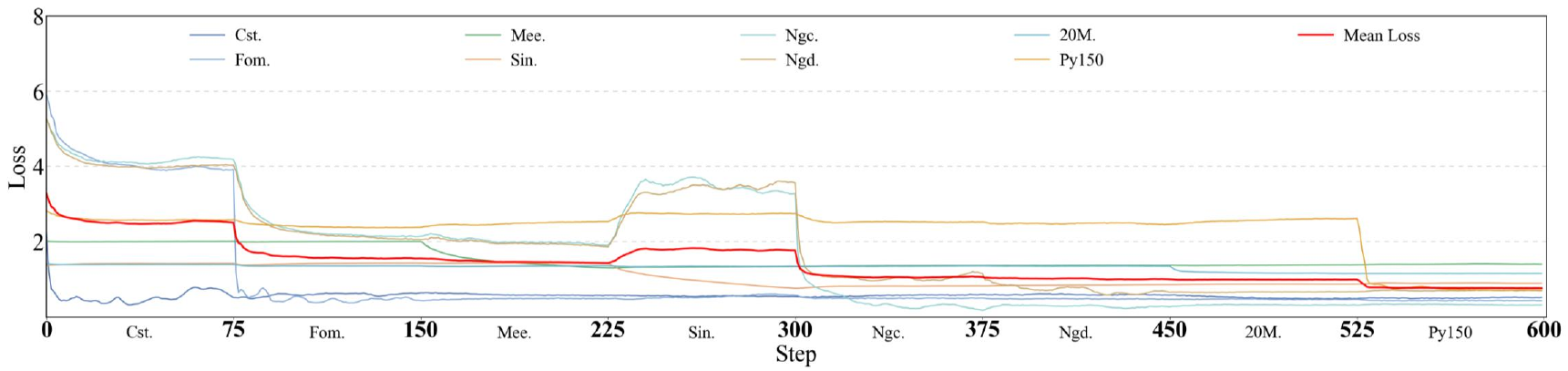}
    \caption{Multi-task Training Loss Curves for Llama3-8B (PIECE-F).}
    \label{fig:llama3_fisher}
\end{figure*}

\begin{figure*}[!t]
    \centering
    \includegraphics[width=\textwidth]{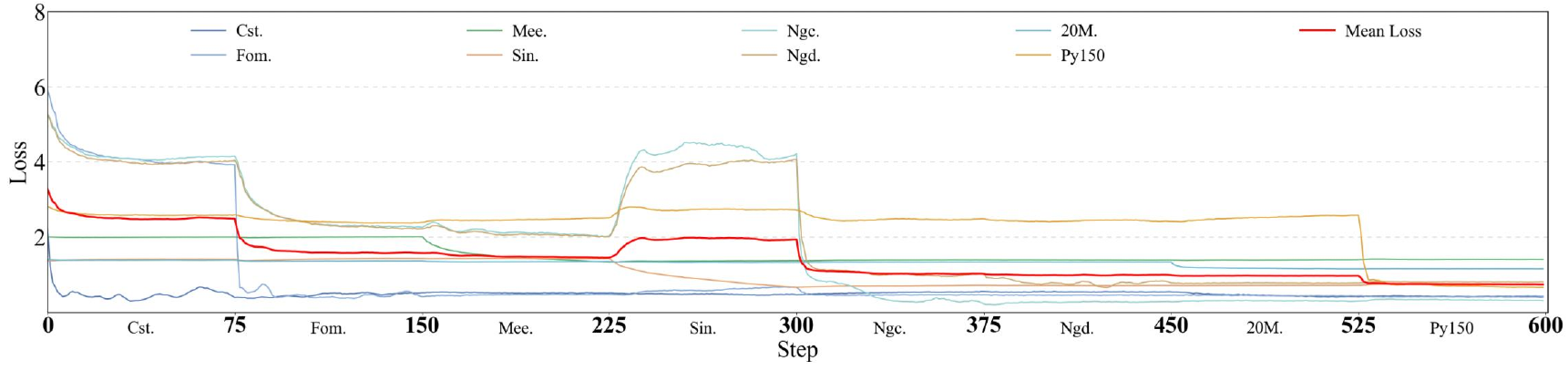}
    \caption{Multi-task Training Loss Curves for Llama3-8B (PIECE-S).}
    \label{fig:llama3_ours}
\end{figure*}

We present several examples of PIECE training logs, as shown in Figures~\ref{fig:llama3_full}-\ref{fig:llama3_ours}, which depict the loss trajectories of individual tasks throughout the continual learning process. To reduce computational overhead, we independently sampled 200 instances from each dataset that were not part of the training or test sets for loss evaluation. The dataset abbreviations on the horizontal axis indicate the task currently being trained.

The results suggest that, compared with SeqFT, PIECE effectively mitigates interference across tasks during training, allowing the model to progressively improve its overall performance in a continual learning setting. Notably, the mathematical reasoning tasks Ngc. and Ngd. appear to be more strongly influenced by other tasks, which may indicate that these tasks impose higher demands on the model’s integrated reasoning capabilities.

\end{document}